\documentclass[12pt]{article}
\usepackage{graphicx,epstopdf,epsfig}
\usepackage{subfigure}
\usepackage{amsmath,amsthm,verbatim,amssymb,setspace}
\usepackage[dvipsnames, svgnames, x11names]{xcolor}
\usepackage{lscape}
\usepackage{threeparttable}
\usepackage{subfigure,multirow,rotating,color}
\usepackage{algorithm, algorithmic}
\usepackage{float}
\usepackage{mathrsfs}
\usepackage{booktabs}
\usepackage{multirow}
\usepackage[hyphens,spaces,obeyspaces]{url}
\usepackage{hyperref}
\usepackage[capitalize]{cleveref}

\usepackage{enumerate}
\usepackage{natbib} 

\newcommand{\blind}{1}

\def\sn{{\rho_{\bx}}}
\def\cH{\mathcal{H}}
\def \cA {\mathcal{A}}

\def\argmin{\mathop{\rm argmin}}

\def\bx{\mathop{\bf x}}

\theoremstyle{plain}

\newtheorem{theorem}{Theorem}[section]

\newtheorem{corollary}[theorem]{Corollary}

\newtheorem{remark}{Remark}
\newtheorem{assumption}[theorem]{Assumption}

\begin{document}

\def\spacingset#1{\renewcommand{\baselinestretch}%
{#1}\small\normalsize} \spacingset{1}


\if1\blind
{
  \title{\bf Transfer Learning for Kernel-based Regression}
   \author{
      Chao Wang$^a\thanks{The first two authors contribute equally to this paper}$, Caixing Wang$^{b*}$, Xin He$^{a\dag}$, Xingdong Feng$^a\thanks{Corresponding authors}$\\
        \\
	$^a$ School of Statistics and Data Science \\
	Shanghai University of Finance and Economics\\
       $^b$ School of Statistics and Data Science \\
	Southeast University\\
    }  
      \date{}
  \maketitle
} \fi

\if0\blind
{
  \bigskip
  \bigskip
  \bigskip
  \begin{center}
    {\LARGE\bf Transfer Learning for Kernel-based Regression}
\end{center}
  \medskip
} \fi

\bigskip

\doublespacing
\begin{abstract}
In recent years, transfer learning has garnered significant attention. Its ability to leverage knowledge from related studies to improve generalization performance in a target study has made it highly appealing. This paper focuses on investigating the transfer learning problem within the context of nonparametric regression over a reproducing kernel Hilbert space. The aim is to bridge the gap between practical effectiveness and theoretical guarantees. We specifically consider two scenarios: one where the transferable sources are known and another where they are unknown. For the known transferable source case, we propose a two-step kernel-based estimator by solely using kernel ridge regression. For the unknown case, we develop a novel method based on an efficient aggregation algorithm, which can automatically detect and alleviate the effects of negative sources. This paper provides the statistical properties of the desired estimators and establishes the minimax rate. Through extensive numerical experiments on synthetic data and real examples, we validate our theoretical findings and demonstrate the effectiveness of our proposed method.
\end{abstract}

\noindent%
\vfill

\newpage
\section{Introduction}
\begin{figure}
    \centering
    \subfigure[Traditional Machine Learning]{
        \includegraphics[width=0.45\textwidth]{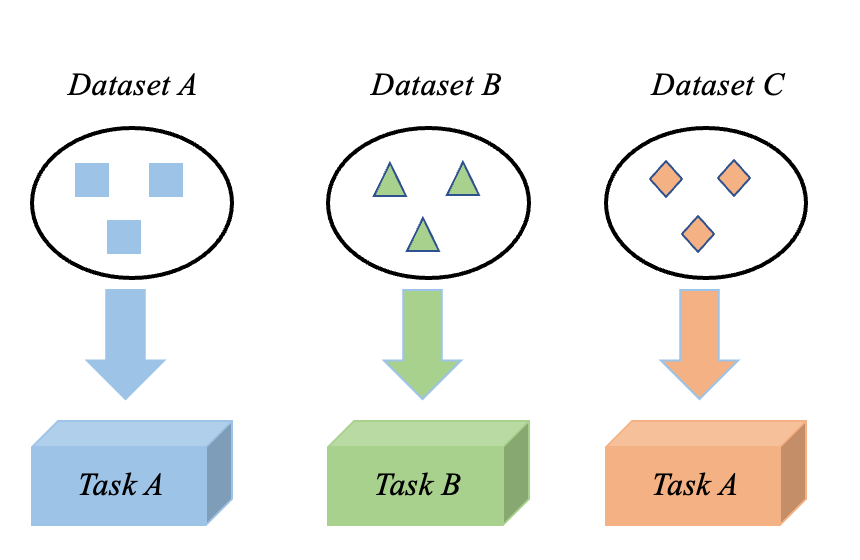}
        \label{pdf_cs}
         }
    \subfigure[Transfer Learning]{
 \includegraphics[width=0.45\textwidth]{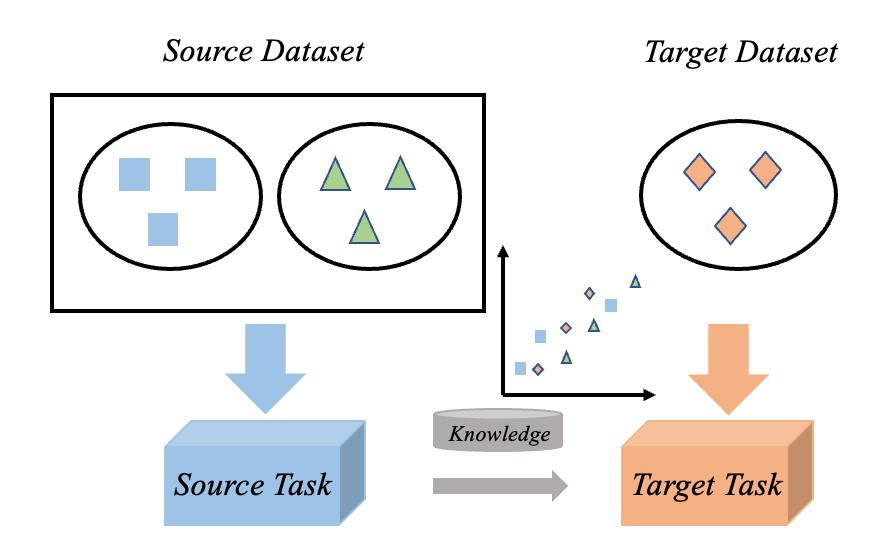}
       
        \label{learning_cs}
    }
  \caption{\footnotesize{Different learning processes between classical machine learning and transfer learning.}}
  \label{fig1}
\end{figure}

Kernel-based nonparametric regression provides an efficient and powerful tool for statistical analysis and has been extensively studied in the fields of machine learning community \citep{kimeldorf1971some,scholkopf2002learning}.  Despite their success, the existing kernel-based methods have mainly focused on the task of learning from the available dataset from one single study, where the sample size can be insufficient in many real applications, and thus weakens the practical performance. Fortunately, with the rapid development of scientific research, many relevant datasets collected from other studies are available and may share some similarities and thus can be used to enhance the learning performance in the target task. In such cases, transfer learning becomes desirable as it allows us to incorporate the knowledge gained from solving other relevant (source) tasks to improve the accuracy of the target task of interest \citep{torrey2010transfer}. Figure \ref{fig1} illustrates the difference between classical machine learning and transfer learning.

Transfer learning has been extensively explored in the literature, boasting a wide array of applications across various domains, including text sentiment classification \citep{do2005transfer}, medical imaging \citep{raghu2019transfusion}, and game playing \citep{taylor2009transfer}. While transfer learning has achieved significant practical success, theoretical studies, particularly in the context of kernel ridge regression (KRR), remain sparse. This paper delves into the transfer learning challenges associated with KRR under the scenario of posterior drift \citep{cai2021transfer,cai2022transfer}, where the conditional distribution of the response given the covariates may change between the source and target tasks, although the covariates' marginal distributions remain consistent. The phenomenon of posterior drift is prevalent in numerous practical scenarios, including crowd-sourcing \citep{zhang2014spectral}, robotics control \citep{vijayakumar2002statistical}, and air quality prediction \citep{wang2016nonparametric}.
It is also interesting to point out that other kinds  of transfer learning problem including covariate shift \citep{shimodaira2000improving,sugiyama2005model} and label shift \citep{garg2020unified,lee2025doubly}, where the former assumes the conditional distribution of the response given the covariates between the source and target is the same while the marginal distributions of the covariates are different, and the latter assumes the conditional distribution of the covariates given the response remains unchanged between the source and target while the marginal distributions of the response differ.
Unless otherwise specified, throughout this paper, transfer learning will refer specifically to the posterior drift setting.

Our primary goal is to address the challenges of nonparametric transfer learning under posterior drift setting from both methodological and theoretical perspectives. Specifically, we concentrate on two prevalent scenarios within transfer learning: (i) scenarios where prior information about transferable sources is available; and (ii) scenarios where such prior information is absent, potentially including the presence of negative sources. For the former scenario, we introduce a two-step, kernel-based estimator that relies exclusively on KRR, demonstrating that this estimator achieves optimality with appropriately selected regularization parameters. In the latter scenario, where some source models may negatively influence the target model, a phenomenon known as a negative source, traditional transfer learning approaches can degrade the predictive performance of the target model \citep{seah2012combating, ge2014handling}. To counter this, we propose an innovative aggregation-based algorithm designed to automatically identify and mitigate the impacts of negative sources. Additionally, we explore the statistical properties of our estimators and establish their minimax lower bounds. We substantiate our theoretical insights and the efficacy of our algorithms through a series of synthetic experiments and real-world applications.

\subsection{Our contributions}
The main contributions of this paper are investigations on nonparametric transfer learning under the posterior drift setting from the aspects of methodology and theory, which is of fundamental interest to the machine learning community. To some extent, this paper fills the gaps between the practical effectiveness and theoretical guarantees of the kernel-based transfer learning regression. It provides the answer to the question of whether the transfer estimation of KRR can achieve optimality and what types of conditions are required. To our limited knowledge, our work is one of the pioneering works to study the posterior drift problem of KRR with solid theoretical guarantees. The specific contributions are concluded as follows.

\begin{itemize}
 \item \textit{Methodology novelty:} 
    We develop a two-step transfer learning algorithm based on the known transferable index set $\cA_h$ and similarity parameter $h$ as defined in \eqref{def_index}, which solely uses KRR and consists of a debiasing step to enhance the transfer learning accuracy. The proposed algorithm is summarized in \cref{Al1} and denoted by $\cA_h$-TKRR.
    For those more challenging scenarios where the transferable sources are unknown,  we develop a novel sparse aggregation-based transfer learning algorithm of KRR in \cref{Al2}, denoted by SA-TKRR, which aggregates multiple estimators and has the ability to automatically eliminate the effects of negative sources.
    \item \textit{Theoretical assessments:} By using the analytic tools for integral operator  \citep{smale2007learning,caponnetto2007optimal}, we establish the upper bound in Theorem \ref{thm_ub_Ah_TKRR}  for $\cA_h$-TKRR 
    and it reveals that with properly chosen regularization parameters, the obtained convergence rate is much faster than that of the standard KRR estimator only using the target data,  provided that $h$ is relatively small.
    More importantly, an algorithm-free minimax lower bound  is also provided in Theorem \ref{theorem_2},  which demonstrates that the convergence rate achieved by  $\cA_h$-TKRR
    is  optimal.
    For a much more challenging case where the transferable source is unknown, we leverage the 
    results in \cite{gaiffas2011hyper} to establish the upper bound for SA-TKRR
    in Theorem \ref{theorem_4}, which indicates that
    SA-TKRR
    can attain
    the best achievable rate for $\cA_h$-TKRR across all values of $h$, up to an additional parametric rate. This confirms the adaptivity of SA-TKRR to the optimal choice of $h$ regardless of the discrepancy between the target and source domains.
    \item \textit{Numerical verification:} Another contribution of this work is the extensive examination of the validity and effectiveness of our proposed algorithms through various synthetic and real-world examples. This empirical evaluation not only supports the theoretical claims made in Theorems \ref{thm_ub_Ah_TKRR} and \ref{theorem_4} but also enhances our understanding of the two methods introduced.
\end{itemize}

\subsection{Related works}
Transfer learning has attracted tremendous interest from both practitioners and researchers, and various transfer learning algorithms have been proposed under different settings \cite{pan2010survey,weiss2016survey}.  Yet, many existing transfer learning methods, especially under the posterior drift setting, are proposed under some specific model assumptions, and the theoretical understanding of those methods is still insufficient and has become increasingly popular.

A majority of the existing methods are developed under the linear model assumption. Specifically, \cite{bastani2021predicting}  studies the transfer learning problem of high-dimensional linear regression and proposes a novel two-step estimator, which efficiently combines the source and target data and provides the upper bound of the estimation error. Yet, it only focuses on the case with one source task and assumes that the sample size of the source data is larger than the dimension of the coefficient. Inspired by this pioneering work, \cite{li2022transfera} considers the problem of multi-source transfer learning in high-dimensional sparse linear regression, and uses the $\ell_q$-distance, $0\leq q \leq 1$, as the similarity measure. They also establish the minimax rates under certain technical conditions.  Moreover, \cite{tian2022transfer} and \cite{li2023estimation} further consider the estimation and inference problem for the high-dimensional generalized linear model with knowledge transfer, and minimax rate of convergence and asymptotic normality for the debiased estimator are established. To model the data heterogeneity, \cite{zhang2022transfer} focuses on the high-dimensional quantile regression problem under the transfer learning framework. Two convolution smoothing transfer learning algorithms are provided and analogous theoretical guarantees are also established.  Note that all the aforementioned methods focus on the linear model case, which is restrictive and difficult to verify in practice.

Recently,  nonparametric transfer learning has also received significant attention.
Specifically, \cite{blitzer2007learning} and \cite{mansour2009domain}  consider the nonparametric domain adaptation in the context of the general classification problem and provide the
uniform convergence bound for the estimator that minimizes a convex combination
of the source and target empirical risk. The problem of nonparametric covariate shift has also attracted considerable attention in recent years \citep{pathak2022new,ma2023optimally,feng2023towards}, where either H{\"o}lder continuous function space or reproducing kernel Hilbert spaces (RKHS) is considered.
Under the posterior drift setting, \cite{cai2021transfer}  proposes a data-driven adaptive classifier for transfer learning of nonparametric classification and conducts a non-asymptotic minimax study, which offers valuable statistical insights and motivates many follow-up works. Inspired by this work, \cite{cai2022transfer} investigates the transfer learning problem of nonparametric regression, where the $\ell_1$-distance is adopted as the similarity measure. Then, a novel confidence thresholding estimator is developed, which can achieve optimality. Yet, it is interesting to point out that they focus on the function class of polynomial functions with  H{\"o}lder continuity, which is still somewhat restrictive and may suffer computational burden, especially when the number of 
covariates is relatively large.  Compared to local polynomial regression, KRR can better handle data with many covariates and is computationally efficient since it has an explicit solution. More importantly, the method in \cite{cai2022transfer} is not able to deal with the presence of  
negative sources, while an efficient nonparametric transfer learning algorithm 
is developed in this paper for mitigating such adverse effects.

\subsection{Paper organization}
The rest of this paper is organized as follows. In Section \ref{sec:meth}, we introduce some necessary notations,
basic knowledge of reproducing kernel Hilbert spaces (RKHS) together with the associated integral operator,
and transfer learning under the kernel-based setting. Section \ref{sec:3} formally formulates the transfer learning problem of KRR and two efficient algorithms are developed to handle the cases where the prior information of transferable sources is known and unknown, respectively.  Section \ref{sec:theory} is devoted to establishing the theoretical guarantees on the proposed algorithms, including the minimax rates. 
 Numerical experiments on synthetic examples and two real applications are provided in Sections \ref{sec:exp} and \ref{real_data}. Section \ref{sec:con} contains a brief conclusion and future directions.  The Supplemental Material contains the extension of the analysis to other similarity measures, additional simulation results, and all the detailed proofs.
 
\section{Preliminaries}\label{sec:meth}
In this section, we introduce some necessary notations,  provide some basic background on reproducing kernel Hilbert spaces, and formulate the problem of transfer learning for the kernel-based nonparametric regression.

\subsection{Notation}
Let $\rho_{\bx}$ be the probability measure on a compact support $\mathcal{X}\subset \mathbb{R}^d$. 
We use ${\cal L}_{2}(\mathcal{X}, \rho_{\bx})=\big\{f:  \int_{{\mathcal{X}}}f^2(\bx)d\rho_{\bx} < \infty\big\}$ to denote the space of square-integrable functions with respect to $\rho_{\bx}$,  equipped with the inner product $\langle f,g \rangle_{\rho_{\bx}}=\int_{\mathcal{X}}f(\bx)g(\bx)d\rho_{\bx}$ and squared norm  $\|f\|^2_{\rho_{\bx}}=\int_{\mathcal{X}}f^2(\bx)d\rho_{\bx}$.  In a slightly abuse of notation, we write $\langle \cdot, \cdot\rangle= \langle \cdot, \cdot\rangle_{\rho_{\bx}}$ for notation simplicity. We use $\operatorname{Tr}(\cdot)$ to denote the trace of an operator or a matrix, and $|{\cal A}|$ to denote the cardinality of a set $\cal A$. For two sequences $\{a_k\}_{k \geq 1}$ and $\{b_k\}_{k \geq 1}$, we write $a_k \lesssim b_k$ or $a_k=O(b_k)$ if $a_k\leq c b_k$ for some universal constant $c>0$, and $a_k \gtrsim b_k$ or $a_k=\Omega(b_k)$ if $a_k\geq c b_k$. We say $a_k$ is of constant order if $a_k=O(1)$. And we say $a_k\asymp b_k$ if both $a_k \lesssim b_k$ and $a_k \gtrsim b_k$ hold. For two random variable sequences $\{A_k\}_{k\ge 1}$ and $\{B_k\}_{k\ge 1}$, we denote $A_k=O_P(B_k)$ if for any $\epsilon>0$, there exists a finite constant $M_\varepsilon>0$ such that  $\sup_k P(|A_k/B_k|>M_\varepsilon)<\epsilon$ and denote $A_k=o_P(B_k)$ if $A_k/B_k\rightarrow 0$ in probability. We use $I$ to represent the identity operator throughout this paper.


\subsection{Reproducing kernel}\label{re_ker}
KRR is one of the most powerful tools in machine learning, where the true target function is often assumed to belong to a reproducing kernel Hilbert space (RKHS),  and has become a time-proven method in the literature of machine learning  \citep{wahba1990spline,murphy2012machine}.

Let ${\cal H}_K$ denote the RKHS induced by a symmetric, positive and semi-definite kernel function $K:{\cal X} \times {\cal X} \rightarrow \mathbb{R}$
and we define its equipped norm as $\|\cdot\|^2_K={\langle \cdot, \cdot\rangle_K}$  with the endowed inner product  $\langle\cdot,\cdot\rangle_K$. We also define $K_{\bx}:=K(\bx,\cdot)\in {\cal H}_K$, for each $\bx \in \mathcal{X}$ and assume that $\sup_{\bx \in {\cal X}} \sqrt{K(\bx, \bx) }\leq \kappa$ for some positive  constant $\kappa$. An important property in $\cH_K$ is called the reproducing property, stating that $
\langle f, K_{\bx}\rangle_K=f(\bx)$
for any 
$ f \in \cH_K.$  
Note that the kernel complexity of $\cH_K$ \citep{Koltchinskii2010SPARSITYIM} is fully characterized  by an integral
operator $L_K: \cH_K \rightarrow \cH_K$, defined as
$L_Kf=\int K_{\bx}f(\bx)d\rho_{\bx}.$
It is well-known that the integral operator $L_K$ is positive, trace-class, self-adjoint, and thus a compact operator.  The spectral theorem implies that   $L_Kf=\sum_{k=1}^{\infty} \mu_{k}  \langle f, \psi_k\rangle \psi_{k}$, where $\mu_{1} \geq \mu_{2} \geq \cdots \geq 0$ is a sequence of nonnegative eigenvalues and $\{\psi_{k}\}_{k=1}^{\infty}$ are the corresponding orthonormal basis in ${\cal L}_{2}\big ({\cal X},\rho_{\bx} \big )$.

\subsection{Multi-source, target model and similarity measure}
We now formulate the transfer learning problem for KRR. 
Suppose that we collect $m+1$ datasets: the target dataset ${\cal T}=\big\{({\bx}^{(0)}_i,y^{(0)}_i)\big\}_{i=1}^{n_0}$ and multi-source datasets ${\cal S}^{(k)}=\big\{({\bx}^{(k)}_{i},y^{(k)}_{i})\big\}_{i=1}^{n_k}$ for  $k=1,\ldots,m$, which are generated by the following mechanism that
\begin{align*}
\big\{({\bx}^{(k)}_i,y^{(k)}_i)\big\}_{i=1}^{n_k} \overset{i.i.d.}{\sim}  \rho^{(k)}_{\bx, y}, \ k=0,1,\ldots,m,
\end{align*}
for some unknown joint distributions $\rho^{(k)}_{\bx, y}$ supported on $\mathcal{X}\times \mathbb{R}$. 
In this paper, we consider the scenario where the sample size of the target data is much smaller than that of
multi-source data due to the difficulties in data collection or high costs. The conditional distribution on $\mathbb{R}$ given $\bx$  is denoted by $\rho^{(k)}_{y|\bx}$. Moreover, we assume that $\rho^{(k)}_{\bx, y}=\rho^{(k)}_{y|\bx}\rho_{\bx},~k=0,\ldots,m$, and then it requires the conditional distributions between the source and target data may be different, while the marginal distributions are the same. Similar requirements are also assumed in the literature of transfer learning \citep{cai2021transfer,cai2022transfer}. Consider a general regression setting that
$$
y_i^{(k)} =   f_\rho^{(k)}({\bx}^{(k)}_i) + \epsilon_i^{(k)},  \ k=0,1,\ldots,m,
$$
where $\epsilon_i^{(k)}$ is a random noise with $E[\epsilon_i^{(k)}|{\bx}^{(k)}_i]=0$ for $i=1,\ldots,n_k$. Thus, there holds $f_\rho^{(k)}({\bx}^{(k)})=\int y^{(k)}d\rho^{(k)}_{y|{\bx}^{(k)}}$, and we further assume  that $f_\rho^{(k)} \in \cH_K$ in this paper.

Our primary interest is to achieve a much more accurate estimation and prediction performance for the KRR by utilizing the target data $\mathcal{T}$ and the useful information provided by the multi-source data $\mathcal{S}^{(k)}$, $k=1,\ldots,m$.  Ideally, if the source function $f_\rho^{(k)}$ is close to the target function $f_\rho^{(0)}$, we can improve the accuracy of estimating $f_\rho^{(0)}$ in the target domain by incorporating information from $\mathcal{S}^{(k)}$ in the auxiliary source domains. Clearly, the effectiveness of such a transfer learning strategy heavily relies on quantifying the similarity between $f_\rho^{(0)}$ and $f_\rho^{(k)}$.  Therefore, we define the $k$-th contrast function as $\delta^{(k)}=f_\rho^{(0)}-f_\rho^{(k)}$ to measure the similarity between the target and source function. Specifically, for  some  small $h>0$, we say the $k$-th source function $f_\rho^{(k)}$ is $h$-transferable if 
\begin{equation}\label{h-tran}
\|\delta^{(k)}\|_K=\|f_\rho^{(0)}-f_\rho^{(k)}\|_K \leq h,
\end{equation}
and then,  the $h$-transferable source index set can be defined as 
\begin{align} \label{def_index}
{\cal A}_h=\big \{k:\|\delta^{(k)}\|_K \leq h, 1\leq k\leq m\big\}.
\end{align}     
It is worth pointing out that the transfer efficiency depends on the choice of similarity level $h$ in that a smaller $h$ may result in a better estimation accuracy by leveraging information from similar transferable source data. Yet, when $h$ shrinks to a near-zero level, it may lead to the case that ${\cal A}_h=\emptyset$ and then all the source data are useless.

Note that the transferable source index set in \eqref{def_index} is defined in terms of RKHS-norm, due to the fact that the RKHS norm of the estimated contrast
functions $\widehat{\delta}^{(k)}$ defined in \eqref{estimated_contrast} can be efficiently computed by the reproducing
property.  We notice that  $\|\cdot\|_\sn$-norm is also a  feasible measure to impose the transferrability
condition, and the 
proposed algorithms in Section \ref{sec:3} can be adapted to the $\|\cdot\|_\sn$-norm with slight modifications.
Details on the modified algorithms as well as the additional theoretical and numerical results are provided in Section S7 of the Supplementary Material.

\section{Transfer Learning for Kernel Ridge Regression}\label{sec:3}
In this section, we propose two efficient transfer learning methods for KRR to deal with those cases where  ${\cal A}_h$  is known and unknown, respectively. For notation simplicity, we  denote $n_{{\cal A}_h}=\sum_{k \in {\cal  A}_h}n_k$, $\eta_k=\frac{n_k}{n_{{\cal  A}_h}+n_0}$ for $k\in \{0\}\cup {\cal  A}_h $, and $m_h=|{\cal  A}_h|+1$. Further 
define  the weighted average function  $f^p_\rho$ as
$  f^p_\rho=\sum_{k\in \{0\}\cup{\cal A}_h}\eta_kf^{(k)}_\rho. $

\begin{algorithm}[t]
	\renewcommand{\algorithmicrequire}{\textbf{Input:}}
	\renewcommand{\algorithmicensure}{\textbf{Output:}}
    \renewcommand{\algorithmicrepeat}{\textbf{Repeat}}
     \renewcommand{\algorithmicuntil}{\textbf{End}}
	\caption{: \textbf{${\cal A}_h$-TKRR}} 
	\begin{algorithmic}[1]\label{Al1}
        \REQUIRE  Multi-source data $\mathcal{S}^{(k)}=\big\{({\bx}^{(k)}_{i},y^{(k)}_{i})\big\}_{i=1}^{n_k}$, $k=1,\ldots,m$,  target data $\mathcal{T}=\big\{({\bx}^{(0)}_i,y^{(0)}_i)\big\}_{i=1}^{n_0}$,  index set ${\cal A}_h$, the pre-specified parameters $\lambda_1$ and $\lambda_2$.
       
	 \STATE \textit{\bf Transferring step}. Compute $\widehat{f}^p$ by solving \eqref{trans_step};
\STATE \textit{\bf Debiasing step}. Compute $\widehat{w}^{(0)}_i=y_i^{(0)}-\widehat{f}^p({\bx}_i^{(0)})$, $i=1,\ldots,n_0$ and obtain  $\widehat{f}^{\text{de}}$ by solving \eqref{de_step};
\STATE Let $\widehat{f}=\widehat{f}^p+\widehat{f}^{\text{de}}$.
\ENSURE The final estimator $\widehat{f}$.
	\end{algorithmic}  
\end{algorithm}

\subsection{Two-step transfer learning with known transferable sources}

Inspired by the ideas in \cite{bastani2021predicting} and \cite{li2022transfera}, when ${\cal A}_h$ is known given some pre-specified $h$,  a two-step transfer learning algorithm named ${\cal A}_h$-TKRR is proposed to leverage the information in ${\cal S}_{{\cal A}_h}=\{{\cal S}^{(k)}:k\in {\cal A}_h\}$. The primary approach involves amalgamating the source datasets ${\cal S}_{{\cal A}_h}$ with the target dataset ${\cal T}$ to derive an initial rough estimator of $f^{(0)}_\rho$, which is formalized by the following optimization problem 
\begin{align}\label{trans_step}
\widehat{f}^p=\argmin_{f \in \cH_K} \Big \{ \frac{1}{n_{{\cal A}_h}+n_0} \sum_{k\in \{0\}\cup{\cal A}_h}\sum_{i=1}^{n_k}\big (f({\bx}^{(k)}_i)-y_i^{(k)}\big )^2+\lambda_1\|f\|_K^2 \Big \}, 
\end{align}
where $\lambda_1$ denotes the regularization parameter. By the representer theorem \citep{kimeldorf1971some}, the minimizer of the optimization task \eqref{trans_step} must have a finite form that 
\begin{align*}
\widehat{f}^p(\bx)=\sum_{k\in \{0\}\cup {\cal A}_h}\sum_{i=1}^{n_k} \widehat{\beta}_i^{(k)} K(\bx, {\bx}_i^{(k)}),    
\end{align*}
where $\{\widehat{\beta}_i^{(k)}\}'s$ are the estimated representer coefficients.  To better understand the transferring step, we define $f_{\lambda_1}^p$ as the minimizer of the expected version of \eqref{trans_step} that
\begin{align}\label{expect_f_p}
f_{\lambda_1}^p=\argmin_{f \in \cH_K} \Big \{  \sum_{k\in \{0\}\cup{\cal A}_h}\frac{n_k}{n_{{\cal A}_h}+n_0} E\big[f({\bx}^{(k)}_i)-y_i^{(k)}\big ]^2+\lambda_1\|f\|_K^2 \Big \}, 
\end{align}
and thus it  satisfies  
$
\sum_{k\in \{0\}\cup{\cal A}_h}\eta_k\int\big (f_{\lambda_1}^p(\bx)-f_{\rho}^{(k)}(\bx)\big )K_{\bx}d\rho_{\bx}+\lambda_1f_{\lambda_1}^p=0
$ by the optimality condition of $f_{\lambda_1}^p$.
Clearly, simple algebra yields $f_{\lambda_1}^p=f_{\rho}^{(0)}- \Delta_{\rho}$, where  $\Delta_{\rho}=(L_K+\lambda_1 I)^{-1}\big (\lambda_1f_{\rho}^{(0)}+L_K(f_\rho^{(0)}-f_\rho^p)\big )$ which denotes the expected version of the bias in the transferring step.

Note that the first term of $\Delta_{\rho}$ is known as the approximation error in literature  {\citep{smale2007learning}}, and it is necessary to debias the second term of $\Delta_{\rho}$, which comes from the transferring procedure. Precisely, the  debiasing step is motivated by the fact that  $(L_K+\lambda_1 I)^{-1}\big (L_K(f_\rho^{(0)}-f_\rho^p)\big )$ is the solution of 
\begin{align*} 
\argmin_{f \in \cH_K} \Big\{E\big[f({\bx}^{(0)}_i)-\big(y_i^{(0)}-f_\rho^p({\bx}_i^{(0)}) \big)\big ]^2+\lambda_1\|f\|_K^2 \Big \}.
\end{align*}
It is thus natural to approximate the transferring bias by fitting KRR on the dataset $\big\{({\bx}^{(0)}_i, y_i^{(0)}-f_\rho^p({\bx}_i^{(0)}))\big\}$.  With $f_\rho^p({\bx}_i^{(0)})$ replaced by its estimate $\widehat{f}^p({\bx}_i^{(0)})$, we compute the residuals  $\widehat{w}^{(0)}_i=y_i^{(0)}-\widehat{f}^p({\bx}_i^{(0)})$, $i=1,\ldots,n_0$, and then obtain a debiased estimator by solving the following optimization task
\begin{align}\label{de_step}
\widehat{f}^{\text{de}}=\argmin _{f \in \cH_K}\Big \{ \frac{1}{n_0} \sum_{i=1}^{n_0}\big (f({\bx}^{(0)}_i)-\widehat{w}_i^{(0)}\big )^2+\lambda_2\|f\|_K^2 \Big \},   
\end{align}
where $\lambda_2$ denotes the regularization parameter. 
Again, by the representer theorem, 
we have 
$\widehat{f}^{\text{de}}(\bx)=\sum_{i=1}^{n_0} \widetilde {\beta}_i^{(0)}  K(\bx, {\bx}_i^{(0)}).    $

Combining the above two estimators, we derive the final estimator of $f^{(0)}_\rho$ that
\begin{align}\label{estimator}
\widehat{f}(\bx)=\widehat{f}^p(\bx)+\widehat{f}^{\text{de}}(\bx)=\sum_{i=1}^{n_0} (\widetilde {\beta}_i^{(0)}+\widehat{\beta}_i^{(0)} )  K(\bx, {\bx}_i^{(0)})+\sum_{k\in {\cal A}_h}\sum_{i=1}^{n_k} \widehat{\beta}_i^{(k)} K(\bx, {\bx}_i^{(k)}).
\end{align}
The proposed two-step transfer learning algorithm is summarized in Algorithm \ref{Al1}.

\subsection{Transfer learning with unknown transferable sources}\label{sec:det}
The success of Algorithm \ref{Al1} requires the prior knowledge of ${\cal A}_h$. In practice, it is unrealistic to know ${\cal A}_h$ in advance, and brute-force transfer can be detrimental to the prediction performance when the source and target data are not closely related \citep{pan2010survey,weiss2016survey}.  In this section, we propose a sparse aggregation-based transfer learning KRR algorithm named SA-TKRR, which is robust to the negative sources.  In sharp contrast to  ${\cal A}_h$-TKRR, which highly depends on the choice of the pre-specified  $h$,  SA-TKRR is able to automatically search the optimal choice of $h$ in an efficient way with a theoretical guarantee. 

Now we introduce SA-TKRR in detail. Specifically, we first randomly and uniformly divide the target data $\mathcal{T}$ into two subsets, $\mathcal{T}_1$ and ${\cal T}_2$.   For notation simplicity, we denote  ${\cal S}^{(0)}={\cal T}_1$. Then, we fit the standard KRR on each $\mathcal{S}^{(k)}$ to obtain the estimators $\widehat{f}^{(k)}$ that  
\begin{align}\label{sec3:standardKRR}
\widehat{f}^{(k)}=\argmin_{f \in \cH_K} \Big \{ \frac{1}{n_{k}} \sum_{i=1}^{n_k}\big (f({\bx}^{(k)}_i)-y_i^{(k)}\big )^2+\lambda^{(k)}\|f\|_K^2 \Big \}, 
\end{align}
for $k=0,1,\ldots,m$. Next, we compute the estimated contrast function 
\begin{align}\label{estimated_contrast}
\widehat{\delta}^{(k)}=\widehat{f}^{(k)}-\widehat{f}^{(0)}.    
\end{align}
Note that the RKHS-norm  $\|\widehat{\delta}^{(k)}\|_K=\|\widehat{f}^{(k)}-\widehat{f}^{(0)}\|_K$ can be directly  computed by using the the representer theorem \citep{kimeldorf1971some} and the property of inner product $\langle\cdot,\cdot\rangle_K$. 
Thus, we  can calculate the estimated rank $\widehat{R}_k$ of $\|\widehat{\delta}^{(k)}\|_K$ in $\{\|\widehat{\delta}^{(1)}\|_K,\ldots,\|\widehat{\delta}^{(m)}\|_K\}$.  It is clear that the smaller $\|\widehat{\delta}^{(k)}\|_K$ is, the closer the $k$-th source function and the target function are. Once the ranks $\{\widehat{R}_k\}_{k=1}^m$ are obtained,  we define an index set that
\begin{align}\label{def:Al}
\widehat{{\cal A}}_{\ell}=\{k: \widehat{R}_k\le \ell,k=1,\ldots,m\},
\end{align}
for $\ell=1,\ldots,m$, and then fit  ${\cal A}_h$-TKRR in Algorithm \ref{Al1} with the target data $\mathcal{T}_1$, source data $\{\mathcal{S}^{(k)}\}_{k \in \widehat{{\cal A}}_{\ell}}$ and parameters $\lambda_{1}^{(\ell)}, \lambda_{2}^{(\ell)}$ to obtain an intermediate  estimator $\widehat{f}_\ell$. Note that we define $ \widehat{{\cal A}}_0=\emptyset$ and thus $\widehat{f}_0=\widehat{f}^{(0)}$.

\begin{algorithm}[t]
\renewcommand{\algorithmicrequire}{\textbf{Input:}}
\renewcommand{\algorithmicensure}{\textbf{Output:}}
\renewcommand{\algorithmicrepeat}{\textbf{Repeat}}
\renewcommand{\algorithmicuntil}{\textbf{End}}
	\caption{\textbf{SA-TKRR}}
	\begin{algorithmic}[1]
        \REQUIRE Multi-source data $\mathcal{S}^{(k)}=\big\{({\bx}^{(k)}_{i},y^{(k)}_{i})\big\}_{i=1}^{n_k}$, $k=1,\ldots,m$,  target data $\mathcal{T}=\big\{({\bx}^{(0)}_i,y^{(0)}_i)\big\}_{i=1}^{n_0}$, the pre-specified parameters $\{\lambda^{(k)}\}_{k\ge0}, \{\lambda_{1}^{(\ell)}\}_{\ell\ge 1}, \{\lambda_{2}^{(\ell)}\}_{\ell\ge1}$, $c$ and $\phi$, 
\STATE Randomly and uniformly split $\mathcal{T}$ into  $\mathcal{T}_1$ and $\mathcal{T}_2$. 
\STATE Fit \eqref{sec3:standardKRR} on each ${\cal S}^{(k)}$ with regularization parameter $\lambda^{(k)}$  for $k=0,1,\ldots,m$  to obtain 
  $\widehat{f}^{(k)}$ where ${\cal S}^{(0)}={\cal T}_1$.
  
\STATE For  $k=1,\ldots,m$, compute the estimated contrast function $\widehat{\delta}^{(k)}$ based on \eqref{estimated_contrast}.
\STATE  For  $k=1,\ldots,m$, compute the rank $\widehat{R}_k$ of $\|\widehat{\delta}^{(k)}\|_K$ in $\{\|\widehat{\delta}^{(k)}\|_K\}_{k=1}^m$.
\STATE  For $ \ell=1,\ldots,m$,  let $ \widehat{{\cal A}}_{\ell}=\{k: \widehat{R}_k\le \ell,1\le k \le m\}$ and obtain $\widehat{f}_\ell$ by  Algorithm \ref{Al1} with $\mathcal{T}_1$ and $\{\mathcal{S}^{(k)}\}_{k \in \widehat{{\cal A}}_{\ell}}$, regularization parameters $\lambda_{1}^{(\ell)}$ and $\lambda_{2}^{(\ell)}$.
 Moreover, let $\widehat{f}_0=\widehat{f}^{(0)}$.
\STATE  Run the hyper-sparse aggregate
algorithm with $\mathcal{F}=\{\widehat{f}_0,\ldots,\widehat{f}_m\}$,  $\mathcal{T}_2$ and parameters  $c, \phi$ to obtain $\widehat{f}_a$ based on \eqref{eq_sparse}. 
 \ENSURE $\widehat{f}_a$.
\end{algorithmic}  \label{Al2}
\end{algorithm}

Then, we obtain a candidate function set ${\cal F}=\{\widehat{f}_0,\widehat{f}_1,\ldots,\widehat{f}_m\}$, and further use an optimal aggregation procedure named the hyper-sparse aggregation algorithm \citep{gaiffas2011hyper} to construct a convex combination $\widetilde{f}$ of the functions in $\cal F$ such that the risk for $\widetilde{f}$ is as close as possible to the minimum risk over $\cal F$. To be more precise, we first randomly and uniformly split ${\cal T}_2$ into two subsets ${\cal T}_{21}$ and ${\cal T}_{22}$. Then, given some pre-specified parameters $\phi$ and $c$, we define a random subset of $\mathcal{F}$ as 
$$     
\widehat{\cal F}_1=\Big \{f\in {\cal F}: \  \widehat{\cal E}_{{\cal T}_{21}}(f)\le \widehat{\cal E}_{{\cal T}_{21}}(\widehat{f}_{{\cal T}_{21}})+c\max \big (\phi \|\widehat{f}_{{\cal T}_{21}}-f\|_{n,1},\phi^2 \big ) \Big \}, 
$$     
where $\widehat{\cal E}_{{\cal T}_{21}}(f)=\frac{1}{|{\cal T}_{21}|}\sum_{(\bx_i,y_i)\in {\cal T}_{21}}(y_i-f({\bx}_i))^2$,  $\widehat{f}_{{\cal T}_{21}}=\argmin_{f\in \mathcal{F}}\widehat{\cal E}_{{\cal T}_{21}}(f)$ and $\|f\|_{n,1}^2=\frac{1}{|{\cal T}_{21}|}\sum_{(\bx_i,y_i)\in {\cal T}_{21}}f^2({\bx}_i)$.    
Moreover, we define  
$$
\widehat{\cal F}_2= \big \{tf_1+(1-t)f_2: \ f_1, \  f_2\in \widehat{\mathcal{F}}_1 \ \text{and}\ 0\le t\le 1 \big \},
$$
which is the collection of convex combinations of at most two functions in $\mathcal{F}$.
Then, the  SA-TKRR estimator $\widehat{f}_a$  can be obtained as 
\begin{align}\label{eq_sparse}
\widehat{f}_a=\argmin_{f \in \widehat{\cal F}_2} \widehat{\cal E}_{{\cal T}_{22}}(f),    
\end{align}
where $\widehat{\cal E}_{{\cal T}_{22}}(f)=\frac{1}{|{\cal T}_{22}|}\sum_{(\bx_i,y_i)\in {\cal T}_{22}}(y_i-f({\bx}_i))^2$.
It is shown in \cite{gaiffas2011hyper} that  \eqref{eq_sparse} has the explicit solution taking of  the form that $\widehat{f}_a=\tilde{t}\widehat{f}_{\ell_1}+(1-\tilde{t})\widehat{f}_{\ell_2}$ where $\ell_1,\ell_2 \in \{0,\ldots, m\}$ and $\ell_1\neq\ell_2$ and the weight $\tilde{t}$ can be directly calculated. It is worth pointing out  that \cite{gaiffas2011hyper} suggest to choose the parameters $\phi$ and $c$  of order $\log(n_0)\log m/n_0$ and  constant order, respectively.

Note that if $\tilde{t}=1$ and ${\ell}_1=0$,  $\widehat{f}_a$ reduces to the standard KRR estimator, which only uses the target data. This ensures that the SA-TKRR estimator can deal with the case where all the source data are negative, and then only the target data are used.  The proposed SA-TKRR algorithm is summarized in Algorithm \ref{Al2}. It is also interesting to notice that the implementation of Algorithm \ref{Al2} does not require the explicit specification of the parameter $h$ as given in \eqref{h-tran}, and can automatically select the best choice of this parameter. In fact, as shown in  Theorem \ref{theorem_4} of Section \ref{sec:theory}, Algorithm \ref{Al2} can automatically select the best threshold $h$ leading to the minimum of the theoretical upper bound. In sharp contrast to most existing algorithms dealing with unknown transferable sources \citep{li2022transfera,tian2022transfer,zhang2022transfer}, the proposed algorithm is much more robust, general and computationally efficient. Its superior performance is also validated by a variety of numerical examples in  Sections \ref{5.2} and \ref{real:2}. It is also interesting to point out that at the end of  Algorithm \ref{Al2}, the estimators $\widehat{f}_{ \ell_1}$ and $\widehat{f}_{\ell_2}$ can be retrained by using the entire target data $\mathcal{T}$ and the corresponding source data $\{\mathcal{S}^{(k)}\}_{k \in \widehat{{\cal A}}_{\ell_1}}$ and $\{\mathcal{S}^{(k)}\}_{k \in \widehat{{\cal A}}_{\ell_2}}$, respectively, to sufficiently utilize the target data.

\begin{remark}
Compared to most existing algorithms dealing with unknown transferable sources \citep{li2022transfera,tian2022transfer,cai2022transfer,zhang2022transfer},   SA-TKRR in Algorithm \ref{Al2} is more robust, general and computationally efficient. Specifically, for robustness, it leverages the hyper-sparse aggregation algorithm \citep{gaiffas2011hyper} and thus is robust against 
the negative sources in the sense that it automatically selects the most appropriate subset to transfer.  Theoretical results in Section \ref{subsec:Theroy_ATL} also demonstrate that SA-TKRR  can automatically adapt to the optimal choice of $h$ and detect transferable source data to enhance the generalization performance in the target domain. In contrast, existing methods can not guarantee the (nearly) optimal performance without additional identification conditions. For generality, we
focus on the RKHS framework, which covers a broad range of function
spaces, and thus is much more general than most existing methods that are restricted to the parametric models. 
For computational efficiency,
the proposed method uses KRR for nonparametric modeling.
KRR can better handle data with many covariates and is computationally efficient as it
has an explicit solution.


\end{remark}

\section{Theoretical Guarantees}\label{sec:theory}
In this section,  we first provide the theoretical guarantee for ${\cal A}_h$-TKRR in \cref{Al1}.
Particularly, a theoretical upper bound of ${\cal A}_h$-TKRR estimator is established in Theorem \ref{thm_ub_Ah_TKRR}, which is also verified as optimal in Theorem \ref{theorem_2}. We also summarize the detailed convergence rates for some special kernel classes. In Section \ref{subsec:Theroy_ATL}, we provide the theoretical results for SA-TKRR in \cref{Al2}
 including the consistency of $\widehat{{\cal A}}_{\ell}$ in Theorem \ref{theorem_3} and the theoretical upper bound of the SA-TKRR estimator $\widehat{f}_a$ in Theorem \ref{theorem_4}. 

We first define a quantity named effective dimension {\citep{caponnetto2007optimal}} to  measure the complexity
of $\cH_K$ with respect to $\rho_{\bx}$ that 
$
{\cal N}(\lambda)=\operatorname{Tr}\big((L_K+\lambda I)^{-1}L_K\big), \ \ \text{for any} \ \lambda>0,
$
which in fact is  the trace of the operator $(L_K+\lambda I)^{-1}L_K$. The following technical assumptions are required to establish the theoretical results.

\begin{assumption}\label{assu1}
Suppose that for each $k=0,1,\ldots,m$,  $E[(y^{(k)})^2] <\infty$   and 
$$
\int(y^{(k)}-f_\rho^{(k)}({\bx}^{(k)}))^b d\rho^{(k)}_{y|{\bx}^{(k)}}\leq \frac{1}{2}b!\gamma^2B^{b-2}, \quad \text{for all}\quad  b \geq 2,
$$
where $B$ and $\gamma$ are some positive constants.  
\end{assumption}

\begin{assumption}\label{assu2}
There exist some constants $\alpha\in (0,1]$ and $Q\ge 1$ such that 
$
{\cal N}(\lambda)\leq Q\lambda^{-\alpha},
$
where $\alpha, Q$ are independent of $\lambda$.
\end{assumption}

\begin{assumption}\label{assu3}
There exists a constant  $r\in[\tfrac{1}{2},1]$   such that $f_{\rho}^{(0)}=L_K^r 
g_{\rho}^{(0)}$ and 
$f_{\rho}^{p}=L_K^r  
g_{\rho}^{p}$ for some  $g_{\rho}^{(0)}, g_{\rho}^{p} \in {\cal L}_2({\cal X}, \rho_{\bx})$,
where $L_K^r$ denotes the $r$-th power of $L_K$.

\end{assumption}

Assumption \ref{assu1}  characterizes the distributions of the noise terms in the source and target models, which is satisfied when the noise term is uniformly bounded, Gaussian or sub-Gaussian distributed. 
Note that a similar condition also appears in \cite{fischer2020sobolev,zhang2023optimality} and it is slightly weaker than the condition on the noise term in the prior works \cite{caponnetto2007optimal,lin2020distributed}.

Assumption \ref{assu2} controls the complexity of the considered $\cH_K$. As pointed out by \cite{lin2020distributed},  when $\alpha=1$, it always holds by taking $Q=\operatorname{Tr}(L_K) \leq \kappa^2$ and when $0 < \alpha < 1$, it is more general than the eigenvalue decaying assumption in literature \citep{caponnetto2007optimal,he2021efficient}.
Assumption \ref{assu3} is a regularity condition on the target regression function $f_{\rho}^{(0)}$ and the weighted average function $f_\rho^p$, where  $r$ controls the smoothness, with a larger $r$ indicating a smoother function. A similar regularity condition is commonly assumed in the literature of kernel methods \cite{smale2007learning,caponnetto2007optimal,fischer2020sobolev}. It is worth pointing out that 
\cref{assu3} is guaranteed by directly 
imposing the smoothness condition
on all regression functions for both target and source domains,  that is, for each $k=0,1,\ldots,m$, 
$f_{\rho}^{(k)}=L_K^rg_{\rho}^{(k)}, \ \text{for some} \ g_{\rho}^{(k)} \in {\cal L}_2({\cal X}, \rho_{\bx})$.

\subsection{Optimal convergence rates for ${\cal A}_h$-TKRR}\label{sub_sec_upper}
The following theorem establishes an upper bound for the ${\cal A}_h$-TKRR estimator.

\begin{theorem}[\textbf{Upper bound of ${\cal A}_h$-TKRR}]\label{thm_ub_Ah_TKRR}
Suppose that Assumptions \ref{assu1}- \ref{assu3} are satisfied and   ${\cal A}_h$ is known with given $h$, with the choices of  $\lambda_1\asymp(n_{{\cal A}_h}+n_0)^{-\frac{1}{2r+\alpha}}$ and $\lambda_2\asymp h^{-\frac{2}{1+\alpha}}n_0^{-\frac{1}{1+\alpha}}$, for any $\delta \in (0,1)$ satisfying  
\begin{align*}
C_0\exp \Big \{-\min\big \{(n_{{\cal A}_h}+n_0)^{\frac{2r+\alpha-1}{2(2r+\alpha)}}(\log(n_{{\cal A}_h}+n_0))^{-\frac{1}{2}},h^{-\frac{2}{1+\alpha}} {{n_0^{\frac{\alpha}{2+2\alpha}}}{(\log n_0)^{-\frac{1}{2}}}}\big \} \Big \} \leq \delta,
\end{align*}
it holds with probability at least $1-\delta$ that
\begin{equation}\label{upp_bound_1}
   \|\widehat{f}-f_{\rho}^{(0)}\|_{\rho_{\bx}} \le  \underbrace{C_1 (n_{{\cal A}_h}+n_0)^{-\frac{r}{2r+\alpha}}\log\frac{12m_h}{\delta}}_{\text{Transferring error}}+\underbrace{C_2h^{\frac{\alpha}{1+\alpha}}n_0^{-\frac{1}{2\alpha+2}}\log\frac{12}{\delta}}_{\text{Debiasing error}},
\end{equation}
where  $C_0, C_1$ and $C_2$ are some universal constants.
\end{theorem}
The proof of Theorem \ref{thm_ub_Ah_TKRR} is provided in Section S4 of the Supplemental Material. The related  constants $\|f_\rho^{(k)}\|_{K},k\in \{0\}\cup \cA_h$  and  $ \|g_\rho^p\|_{\rho_{\bx}}$ are absorbed into the constant $C_1$ in Theorem \ref{thm_ub_Ah_TKRR}, and the  explicit dependence is provided in the proof.
We want to emphasize  that 
the proof of  Theorem \ref{thm_ub_Ah_TKRR} is significantly different from the prior works  \citep{li2022transfera,tian2022transfer,lin2022transfer} where
parametric models are considered. 
Precisely, a novel decoupling method is adopted by constructing an oracle estimator $\widehat{f}^c$ through replacing $\widehat{f}^p$ by the true pooling function ${f}^p_\rho$. Note that ${f}^p_\rho$ is a deterministic quantity in nature. Based on this construction, we develop a novel 
error decomposition in Proposition S4.1 in Section S4 of the Supplemental Material, which successfully decouples the transferring error and the debiased error. More technical details are provided in Section S4 of the Supplemental Material. To the best of our knowledge,  this is a novel treatment to establish the critical theoretical results, highlighting the contribution to a new analytical framework for transfer learning. 

Moreover, the established upper bound in Theorem \ref{thm_ub_Ah_TKRR} consists of two terms 
that are related to the target sample size $n_0$, source sample size $n_{{\cal A}_h}$, and the similarity level $h$. Note that a larger $h$ may lead to a larger size of transferable source data, and thus may further reduce the transferring error at the cost of increasing the debiasing error.  It is also worth noting that when $h$ is sufficiently large the upper bound for $\cA_h$-TKRR is may be larger than that of KRR using the target data alone, which is due to the fact that all the source data with index in $\cA_h$ is used and some sources may differ significantly from the target data.
To be more precise,  the critical threshold on $h$ beyond which leveraging source data is not useful is $\max\{\|f_\rho^{(0)}\|_{K},1\}^{1+\frac{1}{\alpha}}n_0^{-\frac{2r-1}{4r+2\alpha}}$. Note that  the upper bound of 
KRR only using the target data
 is
$\max\{\|f_\rho^{(0)}\|_{K},1\}n_0^{-\frac{r}{2r+\alpha}}$ as shown in \citep{caponnetto2007optimal}.

Moreover, when $h \lesssim n_0^{-\frac{2r-1}{4r+2\alpha}}$ and $n_{{\cal A}_h} \gtrsim  n_0$, the upper bound in \eqref{upp_bound_1} is dominated by the transferring error term $(n_{{\cal A}_h}+n_0)^{-\frac{r}{2r+\alpha}}$, which is much faster than that of KRR. 
When $h=0$ and ${\cal A}_0=\emptyset$, the debiasing error term in \eqref{upp_bound_1} turns to zero and the transferring error terms degenerate to $n_0^{-\frac{r}{2r+\alpha}}$, which implies that no informative source data can be used to improve the estimation accuracy.   Moreover, the upper bound in \eqref{upp_bound_1} becomes sharper as the source data size $n_{{\cal A}_h}$ increases, and then becomes stable if  $(n_{{\cal A}_h}+n_0)^{-\frac{r}{2r+\alpha}}\asymp h^{\frac{\alpha}{1+\alpha}}n_0^{-\frac{1}{2+2\alpha}}$ in the sense that the convergence rate can not be further improved by increasing $n_{{\cal A}_h}$.

\begin{figure}
  \centering
  \includegraphics[scale=0.38]{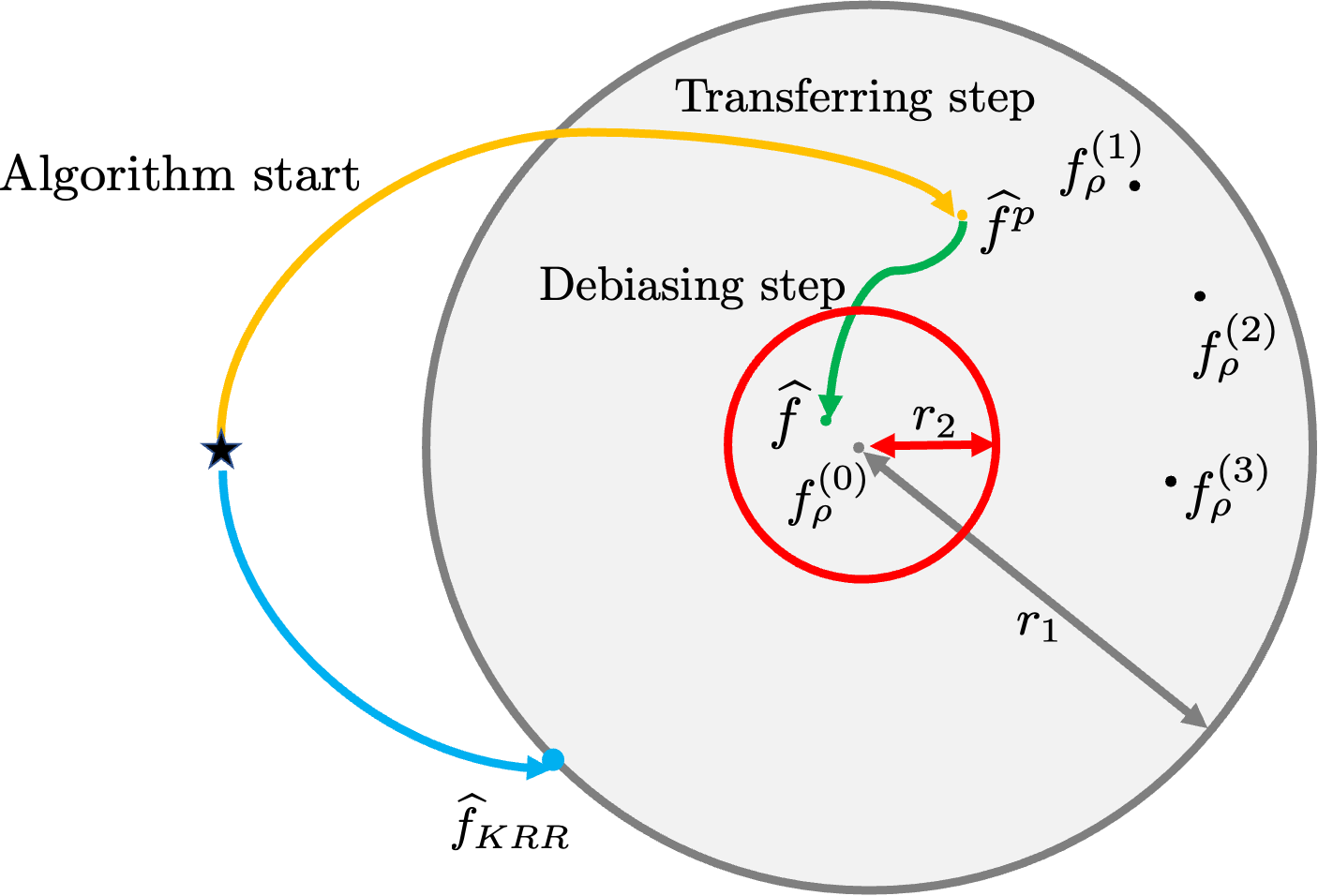}
\caption{\footnotesize{Illustration of how Algorithm \ref{Al1} leads to a better estimator of $f_{\rho}^{(0)}$ in a theoretical example where ${\cal A}_h=\{1,2,3\}$, and $f_{\rho}^{(0)}$,$f_{\rho}^{(1)}$,$f_{\rho}^{(2)}$,$f_{\rho}^{(3)}$ are the corresponding target and  source functions.
}} \label{fig2}
\end{figure}

An example in Figure \ref{fig2} is provided for bringing in some theoretical insights of \cref{Al1}, where $\widehat{f}_{KRR}$ is
the KRR estimator obtained only using the target data. 
It is well-known that the  convergence rate of $\widehat{f}_{KRR}$ is $r_1=O_P(n_0^{-\frac{r}{2r+\alpha}})$ \citep{caponnetto2007optimal}, depicted as the distance from the true regression function  $f_{\rho}^{(0)}$ to the KRR estimator $\widehat{f}_{KRR}$.
In the transferring step, the estimator  $\widehat{f}^p$  is obtained by pooling all the source data indexed by ${\cal A}_h$ and target data. After the debiasing step, we obtain 
the final $\cA_h$-TKRR
estimator $\widehat{f}$. As shown in Theorem \ref{thm_ub_Ah_TKRR}, the upper bound of the  convergence rate of $\widehat{f}$ is $r_2=O_P\big ((n_{{\cal A}_h}+n_0)^{-\frac{r}{2r+\alpha}}+h^{\frac{\alpha}{1+\alpha}}n_0^{-\frac{1}{2+2\alpha}} \big )$ that is much smaller than $r_1$ under mild conditions, so that the proposed 
$\cA_h$-TKRR has a clear advantage over KRR. 

\begin{remark}
We summarize those more explicit convergence rates for some special kernel classes, where two types of kernels are considered as follows:
\noindent (i). {\bf $\beta$-polynomial decay}: 
Polynomial decay kernels have eigen-expansion such that $\mu_j\lesssim j^{-{\beta}}$, which includes the Sobolev and Laplacian kernels. It can be verified that this condition implies Assumption \ref{assu2}  holds with $\alpha=1/\beta$ \citep{caponnetto2007optimal}. Hence, as a straightforward consequence, the convergence rate in Theorem \ref{thm_ub_Ah_TKRR} turns to $O_P \big ((n_{{\cal A}_h}+n_0)^{-\frac{r\beta}{2r\beta+1}}+h^{\frac{1}{1+\beta}}n_0^{-\frac{\beta}{2\beta+2}} \big)$; 
\noindent (ii). {\bf $\gamma$-exponential decay}: 
Exponential decay kernels have eigen-expansion such that $\mu_j\lesssim \exp(-j^{{\gamma}})$. A well-known example is the Gaussian kernel. Interestingly,  Assumption \ref{assu2} holds with an arbitrarily small $\alpha>0$ for the Gaussian kernel, and also leads to the convergence rate in Theorem \ref{thm_ub_Ah_TKRR} holding for arbitrarily small $\alpha>0$. Note that the established results in Theorem~\ref{thm_ub_Ah_TKRR} require $\alpha > 0$, 
and thus exclude the finite-rank kernel case, which corresponds to the parametric transfer learning problem. We want to emphasize that  the problem of parametric transfer learning has been extensively studied 
\citep{li2022transfera,li2022transferb,zhang2022transfer}, the primary contribution of this paper is to provide efficient and useful algorithms for the nonparametric transfer learning problem under the posterior drift setting from the aspects of both methodology and theory.
\end{remark}

Now we turn to establish the lower bounds for all learning methods and show that the ${\cal A}_h$-TKRR estimator is (nearly) optimal.  For notation simplicity, we denote ${\cal Z}^{n_0+n_{{\cal A}_h}}=\big \{{{({\bx}^{(k)}_i,y^{(k)}_i)}_{i=1}^{n_k}}\big\}_{k\in \{0\} \cup {\cal A}_h}$   as the data consisting of the target data and all the transferable source data. 
Moreover, we consider the set of joint probability distributions on ${\cal Z}^{n_0+n_{{\cal A}_h}}$ that
${\cal P}_h=\big\{\otimes_{k\in \{0\} \cup {\cal A}_h} {\rho_{\bx,y}^{(k)}}^{\otimes n_k}\big\}$, where $\otimes_{k=1}^J\rho_{\bx,y}^{(k)}=\rho_{\bx,y}^{(1)}\otimes\cdots\otimes \rho_{\bx,y}^{(J)}$ represents the product of probability distributions and $ {\rho_{\bx,y}^{(k)}}^{\otimes n_k}=\otimes_{i=1}^{n_k}{\rho_{\bx,y}^{(k)}}$. 
\begin{theorem}[\textbf{Minimax lower bound}]\label{theorem_2}
Given ${\cal A}_h$ and  let $\mathcal{P}_0\subset \mathcal{P}_h$ consist of all the distributions satisfying  Assumptions \ref{assu1}-\ref{assu3}. Then,  for any 
$\delta\in (0,1)$ and 
any estimator $\widehat{f}_{0,h}$ based on the data ${\cal Z}^{n_0+n_{{\cal A}_h}}$,  there exists a joint probability distribution in $\mathcal{P}_0$ from which ${\cal Z}^{n_0+n_{{\cal A}_h}}$ is generated such that with probability at least $1-\delta$, 
\begin{align*}
\|\widehat{f}_{0,h}-f_\rho^{(0)}\|_{\rho_{\bx}}^2\ge  C_3\delta \Psi_n,
\end{align*}
where  $C_3$ is some universal constant and 
$$
\Psi_n= \left\{
\begin{array}{lc}
(n_{{\cal A}_h}+n_0)^{-\frac{2r}{2r+\alpha}}+h^{\frac{2\alpha}{1+\alpha}}n_0^{-\frac{1}{1+\alpha}},  & 0<\alpha<1; \\
(n_{{\cal A}_h}+n_0)^{-\frac{2r}{2r+1}}(\log (n_{{\cal A}_h}+n_0))^{-2}+hn_0^{-\frac{1}{2}} (\log n_0)^{-2}, &  \alpha=1 .\\
\end{array}
\right. 
$$
\end{theorem}

Note that for any given $h$,  the infimum in the lower bound  of  Theorem \ref{theorem_2} is taken over all estimators constructed from ${\cal Z}^{n_0+n_{{\cal A}_h}}$. Similarly, the established upper bound for $\cA_h$-TKRR in Theorem \ref{thm_ub_Ah_TKRR}  holds for any given $h$. Given the same  $h$ in Theorems \ref{thm_ub_Ah_TKRR} and \ref{theorem_2},  comparing the upper and lower bounds can conclude that the proposed estimator ${\cal A}_h$-TKRR is exactly optimal under the case  $0<\alpha<1$ and nearly optimal up to $\log$ terms under the case $\alpha=1$. We also notice that the minimax lower bound reduces to the known lower bound
$n^{-\frac{2r}{2r+\alpha}}$ if  $h=0$ and $\mathcal{A}_0=\emptyset$, which concurs with the existing optimal result of the standard KRR \citep{caponnetto2007optimal}.

\subsection{Optimal convergence rates for SA-TKRR}\label{subsec:Theroy_ATL}
Although $\cA_h$-TKRR can  achieve the minimax lower bound in \cref{theorem_2}, it depends on the parameter $h$ and the prior knowledge of $\cA_h$—which are typically unavailable
in practice. To address this limitation, SA-TKRR is proposed in \cref{Al2}, which frees the dependency on $h$ and the prior knowledge of
$\cA_h$
regardless of the discrepancy between the target and source domains.
In this section, we investigate 
the theoretical properties of  SA-TKRR.

To establish the theoretical guarantee,  we  define $h_\ell$ as the smallest value such that  $|{\cal A}_{h_\ell}|\ge\ell$ for  $ 0\le \ell \le m$, where $
{\cal A}_{h_\ell}=\big \{k:\|\delta^{(k)}\|_K \leq h_\ell, 1\leq k\leq m\big\}$, and denote $e_\ell={|{\cal A}_{h_\ell}|}\in \{0,1,\ldots,m\}.$  It is worth pointing out that the cardinality $e_{\ell}$ is unknown in practice and it is introduced for theoretical simplicity.
Note that
by definition, $h_0=0$  and  ${\cal A}_{h_0}=\{k :  \|\delta^{(k)}\|_K=0, 1\le k\le m\}$, and there may also exist some $0\le i  \neq j\le m$  such that $h_i=h_j$ and $e_i=e_j$ due to the fact that there may exist duplicated values in the set $\{0, \|\delta^{(1)}\|_K,\ldots,\|\delta^{(m)}\|_K\}$. 
For instance, suppose that we have three source samples where $\|\delta^{(1)}\|_K= 0$ and $ \|\delta^{(2)}\|_K=\|\delta^{(3)}\|_K$. Clearly, we have $h_0=h_1=0$  with ${\cal A}_{h_0}={\cal A}_{h_1}=\{1\}$, and  $h_2=h_3$ with ${\cal A}_{h_2}={\cal A}_{h_3}=\{1,2,3\}$, and consequently, we have $e_0=e_1=1,e_2=e_3=3$.  
Hence,  we can not establish the consistency result between   $\widehat{\cal A}_\ell$ and ${\cal A}_{h_\ell}$ due to the identifiable issue.
  Alternatively,   for each $\ell$, we denote $\widehat{{\cal A}}_{e_\ell}=\{k: \widehat{R}_k\le e_\ell, k=1,\ldots,m \}$, and there holds that  $\widehat{{\cal A}}_{e_\ell} \in \{\widehat{{\cal A}}_{0},\widehat{{\cal A}}_{1},\ldots, \widehat{{\cal A}}_{m}\}$. Then, we turn to derive the consistency for some sets in  $\{\widehat{{\cal A}}_{0}, \ldots, \widehat{{\cal A}}_{m}\}$, which is sufficient to ensure the nearly optimal convergence rate of SA-TKRR established in Theorem \ref{theorem_4}. 
  In addition to the smoothness condition on $f_\rho^{(0)}$ and $f_\rho^p$ in \cref{assu3},
establishing the consistency of  $\widehat{{\cal A}}_{e_\ell}$ for each $\ell$ requires the 
following smoothness assumption on all regression functions for target and
source domains.
  
\begin{assumption}\label{assu3_exten}
There exists a constant $r\in(\tfrac{1}{2},1]$ such that for each 
$k=0,1,\cdots,m$,
$f_{\rho}^{(k)}=L_K^r
g_{\rho}^{(k)}$  for some  $g_{\rho}^{(k)}\in {\cal L}_2({\cal X}, \rho_{\bx})$.
\end{assumption}
It is worth pointing out that the analysis of $\cA_h$-TKRR only requires \cref{assu3}, while \cref{assu3_exten} is further needed for SA-TKRR in order to ensure the consistency of $\widehat{{\cal A}}_{e_\ell}$ for each $\ell$. Note that \cref{assu3_exten} implies that \cref{assu3} holds with $r$ in \cref{assu3_exten}. 
The following theorem provides consistency property of  $\widehat{{\cal A}}_{e_\ell}$ for each $\ell$.

\begin{theorem}[\textbf{Consistency of $\widehat{{\cal A}}_{e_\ell}$}]\label{theorem_3}
Suppose that  Assumptions \ref{assu1}, \ref{assu2} and \ref{assu3_exten} are satisfied.
Then, with the choices of $\lambda^{(k)}\asymp n_k^{-\frac{1}{2r+\alpha}}$ for $0\le k \le m$,  as $\min_{0\le k\le m} {n_k}\rightarrow\infty$,  we have
$P(\widehat{{\cal A}}_{e_\ell}={\cal A}_{h_\ell})\rightarrow 1.$
\end{theorem}

Theorem \ref{theorem_3} shows that the index set $\widehat{{\cal A}}_{e_\ell}$  estimated by  SA-TKRR equals to the true ``$h_{\ell}$-transferable''  set ${\cal A}_{h_{\ell}}$ with probability tending to 1, which is crucial to establish the convergence rate of SA-TKRR.  In contrast to many existing methods \citep{li2022transfera,tian2022transfer,lin2022transfer,zhang2022transfer}, SA-TKRR  impose no additional identifiable assumptions to ensure the consistency of $\widehat{{\cal A}}_{e_\ell}$, due to the fact that it holds  $\max_{k\in{\cal A}_{h_\ell} }\|\delta^{(k)}\|_K\leq {h_\ell}< \min_{k\not \in{\cal A}_{h_\ell} }\|\delta^{(k)}\|_K$ straightforwardly by the definition of ${\cal A}_{h_\ell}$. 
If we further assume that the elements of  $\{0, \|\delta^{(1)}\|_K,....,\|\delta^{(m)}\|_K\}$ are distinct, we can establish the consistency result of  $\widehat{{\cal A}}_{\ell}$ defined in \eqref{def:Al}.

\begin{corollary}[\textbf{Consistency of $\widehat{{\cal A}}_{\ell}$}]\label{cor:1}
 Suppose that all the assumptions in Theorem \ref{theorem_3} are satisfied, and   $\|\delta^{(k)}\|_K>0$ for $1\le k\le m$ and $\|\delta^{(i)}\|_K\not = \|\delta^{(j)}\|_K$ for all $1\le i,j\le m$ and $i\not= j$.  Then, $h_\ell$ turns to the smallest value such that $|\mathcal{A}_{h_\ell}|=\ell$ and we have
$P(\widehat{{\cal A}}_{\ell}={\cal A}_{h_\ell})\rightarrow 1 $ for each $\ell$. 
\end{corollary}

Now with propositions provided in Section S6 of the Supplemental Material and Theorem \ref{theorem_3}, we are ready to establish the convergence rate of the SA-TKRR estimator.
\begin{theorem}[\textbf{Upper bound of SA-TKRR}]\label{theorem_4}
Suppose that  Assumptions \ref{assu1}, \ref{assu2} and \ref{assu3_exten} are satisfied.
Then, with the choices of  $\lambda^{(k)}\asymp n_k^{-\frac{1}{2r'+\alpha}}$, $\lambda_1^{(\ell)}\asymp(n_{{\cal A}_{h_\ell}}+n_0)^{-\frac{1}{2r+\alpha}}$, $\lambda_2^{(\ell)}\asymp {h_\ell}^{-\frac{2}{1+\alpha}}n_0^{-\frac{1}{1+\alpha}}$, $\phi\asymp \sqrt{\frac{\log m \log n_0} {n_0}}$ and $c=O(1)$, we have
 \begin{align*}
 \|\widehat{f}_a-f^{(0)}_\rho\|_{\rho_{\bx}}^2= O_P\left(\inf_{h\ge 0}\left\{(n_{{\cal A}_h}+n_0)^{-\frac{2r}{2r+\alpha}}+h^{\frac{2\alpha}{1+\alpha}}n_0^{-\frac{1}{1+\alpha}}\right\}+\frac{\log m \log n_0}{n_0}\right).
\end{align*}
\end{theorem}


The established convergence rate of SA-TKRR in Theorem \ref{theorem_4}  involves two components.
The first corresponds to the best achievable rate of  ${\cal A}_h$-TKRR across all values of
$h$, while 
 the second accounts for the error of performing 
 aggregation in SA-TKRR.
The latter is of order ${\log m \log n_0}/{n_0}$,
 which is asymptotically negligible compared to the dominant first term, up to $\log$ factors.
It is also worth mentioning that
$\cA_h$-TKRR  may have degraded performance 
compared to the standard KRR using the target data alone when $h$ is sufficiently large since some sources may differ significantly from the target data.
In contrast, the convergence rate in \cref{theorem_4} is at least faster than $O_P((n_{{\cal A}_0}+n_0)^{-\frac{2r}{2r+\alpha}}+{\log m \log n_0}/{n_0})$, aligning with the 
convergence rate of the standard KRR using the data $\{\mathcal{S}^{(k)}\}_{k\in \{0\}\cup {\cal A}_0}$ plus an additional negligible aggregation error. We also note that the first part $(n_{{\cal A}_0}+n_0)^{-\frac{2r}{2r+\alpha}}$ is faster than $n_0^{-\frac{2r}{2r+\alpha}}$, and degrades to $n_0^{-\frac{2r}{2r+\alpha}}$ when ${\cal A}_0=\emptyset$. Overall, SA-TKRR can automatically adapt to the optimal choice of $h$ and detect transferable source data to enhance the generalization performance in the target domain. 
It is also worth pointing that in Theorem \ref{theorem_4}, the choice of parameters $\lambda^{(\ell)}_1$ and $\lambda^{(\ell)}_2$ depends on the oracle quantity ${\cal A}_{h_{\ell}}$. This choice is primarily used from the perspective of theoretical analysis and in practice, some data-driven strategy, such as cross-validation, can be applied to choose $\lambda^{(\ell)}_1$ and $\lambda^{(\ell)}_2$.

\section{Simulated Experiments}\label{sec:exp}
In this section, we validate our theoretical findings by applying the proposed method and some other competitors to several synthetic examples.   Specifically,  we compare the numerical performance of ${\cal A}_h$-TKRR to some state-of-the-art methods in Section \ref{5.1} where ${\cal A}_h$ is known, and verify our theoretical analysis established in  Section \ref{sub_sec_upper} by varying $n_{{\cal A}_h}$. Additional experiments to investigate the effect of $h$ and $n_0$ is provided in Section S1 of the Supplementary Materials. In Section \ref{5.2}, we also compare the performance of  SA-TKRR  to some state-of-the-art methods where  ${\cal A}_h$ is unknown. In all the numerical examples, we consider the RKHS $\cH_K$ induced by the Gaussian kernel {$K(\bx, {\bx}^\prime)=\exp(-\|\bx-{\bx}^\prime\|^2)$} and
the regularization  parameters $\lambda_1,\lambda_2, \lambda_1^{(\ell)}$ and 
$\lambda_2^{(\ell)}$ 
are selected via cross-validation. 
%

\subsection{Transfer KRR when ${\cal A}_h$ is known }\label{5.1}

In this part, we investigate the performance of ${\cal A}_h$-TKRR  and its competitors, including  KRR where only the target data is used, and ${\cal A}_h$-TKRR-WD where Algorithm \ref{Al1} is implemented without the debiasing step.  The following three synthetic examples are considered, where the number of sources is set as $m=10$.

\noindent\textbf{Example 1.} We first generate $x^{(k)}_i \in \mathbb{R}$ independently from $ U(0,1)$ where $k=0,\ldots,10$ and $i=1,\ldots,n_k$. The responses $y_i^{(k)}$ are independently generated as $
y_i^{(k)} =   f_\rho^{(k)}(x^{(k)}_i) + \epsilon_i^{(k)}$,  where $\epsilon_i^{(k)}\sim N(0,0.4^2)$ and
$f^{(k)}_\rho(x_i^{(k)})=3\sin(3\pi x_i^{(k)})-1.5e^{|x_i^{(k)}-s^{(k)}-0.5|}$,
with $s^{(0)}=0$ and $s^{(k)}$. Here  $ s^{(k)}$'s are independently drawn from  $U(0,s)$ with  some pre-specified parameter $s$. Note that $s$ can be treated as a similarity parameter that measures the difference between the target model and the source models.

\noindent\textbf{Example 2.} The generating scheme is the same as Example 1 except that ${\bx}_i^{(k)}=(x^{(k)}_{1i},x^{(k)}_{2i},\\
x^{(k)}_{3i})^\top\in \mathbb{R}^3$, $f^{(k)}_\rho$ has the form that 
$f^{(k)}_\rho({\bx}_i^{(k)})=\sin(3\pi x_{1i}^{(k)})+3 |x^{(k)}_{1i}-s^{(k)}-0.5|- e^{(x^{(k)}_{2i})^2-(x^{(k)}_{3i})^2},  $
and $\epsilon_i^{(k)}\sim N(0,0.3^2)$.

\noindent\textbf{Example 3.}  The generating scheme is exactly the same as Example 2 except that ${\bx}_i^{(k)}=(x^{(k)}_{1i},\ldots,x^{(k)}_{10 i})^\top\in \mathbb{R}^{10}$ and $f^{(k)}_\rho$ has the form that 
$f^{(k)}_\rho({\bx}_i^{(k)})=\sin(0.75\pi w_{1i}^{(k)})+4 |w_{2i}^{(k)}-s^{(k)}-0.5|- e^{w_{3i}^{(k)}},  $
with $w_{1i}^{(k)}=x^{(k)}_{1i}+x^{(k)}_{4i}+x^{(k)}_{5i}+x^{(k)}_{6i}, w_{2i}^{(k)}=(x^{(k)}_{1i}+x^{(k)}_{2i}+x^{(k)}_{3i})/3, w_{3i}^{(k)}=(x^{(k)}_{7i})^2+(x^{(k)}_{8i})^2-(x^{(k)}_{9i})^2-(x^{(k)}_{10i})^2$. 

\begin{figure}[ht]
    \centering
\includegraphics[width=\textwidth]{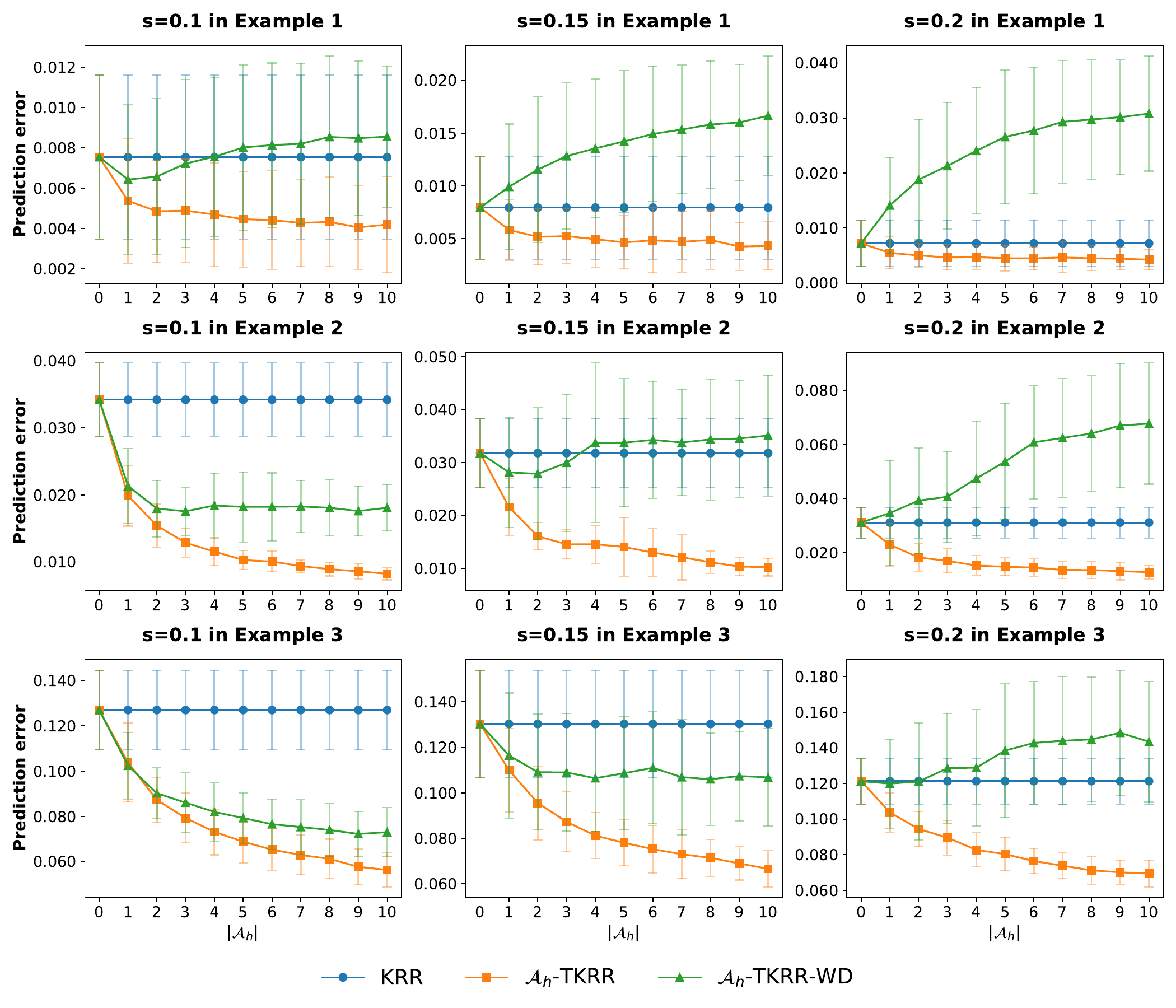}
    \caption{Averaged prediction errors of KRR, 
    $\cA_h$-TKRR and $\cA_h$-TKRR-WD in Examples 1–3 under various scenarios with varying $|\mathcal{A}_h|$.}
    \label{MSE_h_m}
\end{figure}

It is thus clear that the scalar covariate is used in Example 1, and the multi-dimensional covariates are considered in Examples 2 and 3. In all the numerical experiments, unless specified otherwise, we set $n_0=200$ and $n_k=150$ for each $k>0$ in Example 1 and set $n_0=600$ and $n_k=300$ for each $k>0$ in both Examples 2 and 3. Moreover,  we assume that all the sources are known to be transferable in the sense that ${\cal A}_h=\{1,2,\ldots,10\}$. We investigate how the performance of ${\cal A}_h$-TKRR is affected by the number of sources. Specifically, we apply all the competitors to Examples 1--3 and vary 
$|{\cal A}_h|$  from $\{0,1,\ldots,10\}$  in the sense that ${\cal A}_h=\{1,\ldots,|{\cal A}_h|\}$ for $|{\cal A}_h|>0$ and  $s$ from $\{0.10,0.15,0.20\}$.  Each of the scenarios is replicated 100 times and the averaged performance of all the competitors is evaluated by the prediction error of the estimator $\widehat{f}$ that $\|\widehat{f}-f_\rho\|_{n_{te}} ^2=\frac{1}{n_{te}}\sum_{i=1}^{n_{te}} (\widehat{f}({\bx}_i)-f_\rho^{(0)}({\bx}_i))^2$ with a new testing dataset drawn from the target model with $n_{te}=500$.  The results are summarized in Figure \ref{MSE_h_m}. 

From Figure \ref{MSE_h_m}, we observe that ${\cal A}_h$-TKRR outperforms  KRR and ${\cal A}_h$-TKRR-WD in all the scenarios where $|{\cal A}_h|>0$. This indicates that the source data can contribute to the
transfer learning of the target model, and the debiased step further improves the learning accuracy.
Moreover, in all the scenarios, the prediction error of ${\cal A}_h$-TKRR decreases as $|{\cal A}_h|$ grows at first, and then flattens out, which closely coincides with the results in Theorem \ref{thm_ub_Ah_TKRR}.  It is also worth noting that for the larger $s$, the curves of  ${\cal A}_h$-TKRR flatten out earlier, which is also consistent with our theoretical findings.

\subsection{Transfer KRR when {${\cal A}_h$} is unknown}\label{5.2}
In Section \ref{5.1}, we assume that the transferable sources are known, yet such prior information may not be available in practice. In this part, we evaluate the numerical performance of  SA-TKRR proposed in Section \ref{sec:det} and compare it to some state-of-the-art competitors, including KRR where only the target data are used, Pooled-TKRR where Algorithm \ref{Al1} is implemented with all sources, 
D-TKRR where Algorithm \ref{Al1} is implemented with the detection procedure used in \cite{tian2022transfer} and AEW-TKRR, where Algorithm \ref{Al2} is implemented using exponential weight aggregation \citep{leung2006information} without the re-train step.

\begin{figure}[ht]
    \centering
\includegraphics[width=\textwidth]{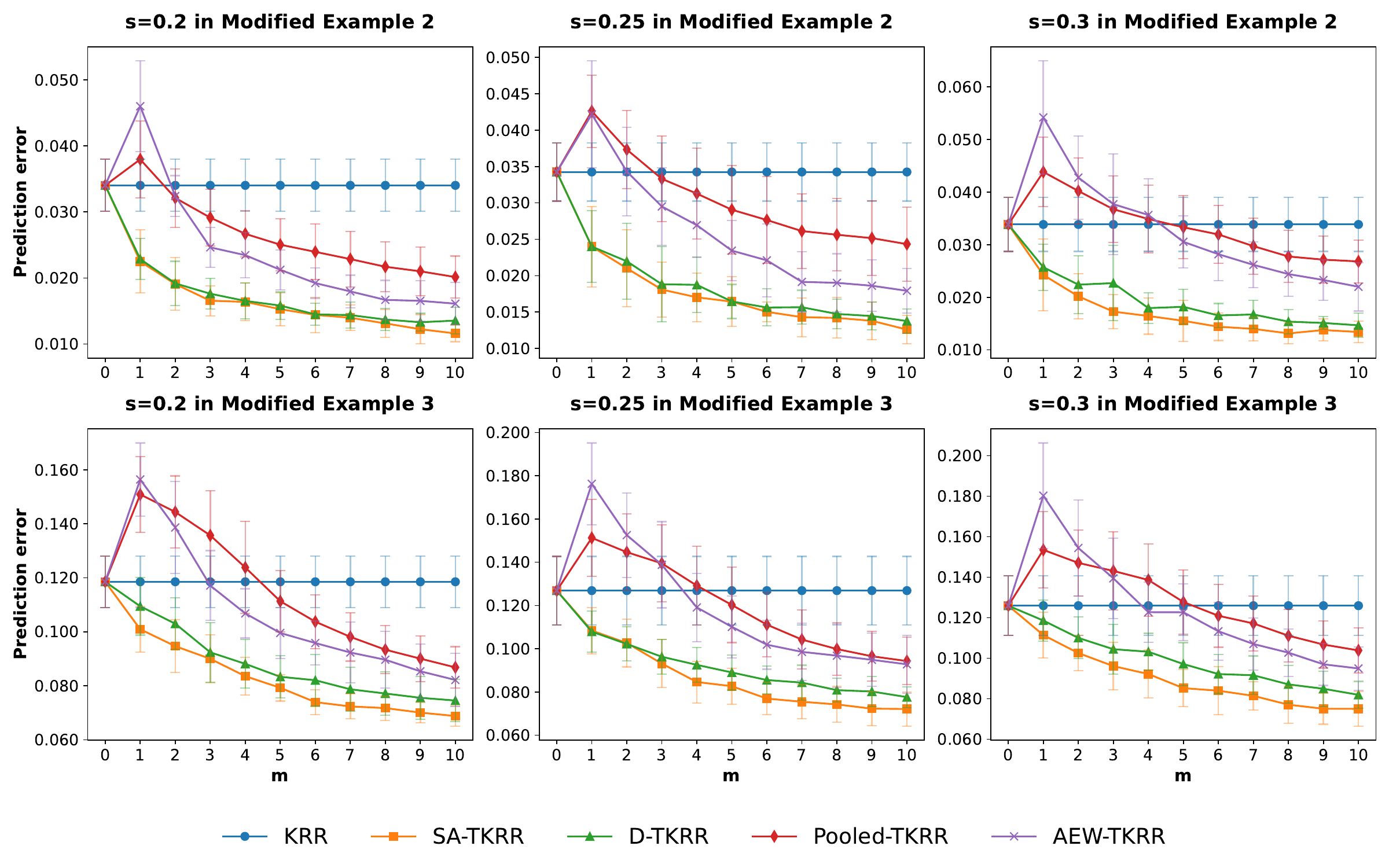}
    \caption{Averaged prediction errors of all the competitors in modified Examples 2 and 3 under various scenarios
with varying $m$.}
\label{MSE_detection}
\end{figure}

Specifically, we consider the multi-dimensional cases and assume that the number of sources is $m+3$. Specifically, the data generating schemes are the same as Examples 2 and 3 in Section \ref{5.1} except that for the last 3 sources,  $s^{(m+1)}, s^{(m+2)}, s^{(m+3)}$ are independently drawn from $U(s,0.4)$, and we denote these examples as modified Example 2 and 3, respectively.  Note that ${\cal A}_h$ is unknown, and thus we apply  SA-TKRR and all the other competitors to Examples 2 and 3 and vary $m$ from $\{0,1,2,\ldots,10\}$ and $s$ from $\{0.2, 0.25, 0.3\}$.  Each of the scenarios is replicated 100 times and the averaged performance of all the competitors is
evaluated by the prediction error presented in Figure \ref{MSE_detection}.

It is thus clear from Figure \ref{MSE_detection} that  SA-TKRR outperforms all the other competitors in almost all the scenarios, which indicates the effectiveness of SA-TKRR in not only capturing useful information contained in the source data but also mitigating the impact of negative sources. This observation also supports our theoretical findings in Section \ref{subsec:Theroy_ATL}. It is worth pointing out that the prediction performance of Pooled-TKRR is even worse than  KRR in most settings,  which indicates that simple transfer learning with all the source data is not preferable since the existing negative sources can bring a harmful impact. The performance of 
AEW-TKRR is less satisfactory, especially when $m$ is relatively small, which is largely due to its required target data splitting procedure.  Note that although the performance of D-TKRR is close to that of SA-TKRR, it is much more computationally challenging due to the adopted detection procedure. It is also interesting to point out that all of the numerical analysis are conducted by using fixed $\phi$ and $c$ following
the theoretical suggestion in Section \ref{subsec:Theroy_ATL}, and it leads to a satisfactory numerical result. The data-driven selection procedure may further improve the numerical performance, but at the cost of
increased computational burden.

\section{Real Data Analysis}\label{real_data}
In this section, we apply the proposed algorithms to two real applications, where we consider the single source case with known ${\cal A}_h$ in the first application and the multiple source case with unknown ${\cal A}_h$ in the second application. In both examples,  we follow the same treatment as those in Section \ref{sec:exp} for the choices of RKHS, tuning parameters,  
and we report the prediction error of the response on the test data, where $f_\rho^{(0)}({\bx}_i)$ is replaced by $y_i$ such that $\frac{1}{n_{te}}\sum_{i=1}^{n_{te}} (\widehat{f}({\bx}_i)-y_i)^2$.

\subsection{Wine quality data}
In this subsection, we consider a real application to the Portuguese Vinho Verde wine data \citep{cortez2009modeling}, which is available in \url{https://archive.ics.uci.edu/dataset/186/wine+quality}.  This dataset contains the qualities of red and white wines and consists of the response (wine quality) and 13 covariates, including fixed acidity, citric acid and alcohol.  Note that the two types of wines share the same covariates, but the underlying model may not be the same, and thus it is reasonable to consider transfer learning.  Following the same treatment as that in \cite{cai2022transfer}, we use the white wine samples as the source data and the red wine samples as the target data, and the sample sizes are  4898 and 1599, respectively. Then, the primary interest is to enhance the prediction accuracy of the red wine quality with the help of the white wine data.

\begin{figure}[ht]
    \centering
\includegraphics[width=\textwidth]{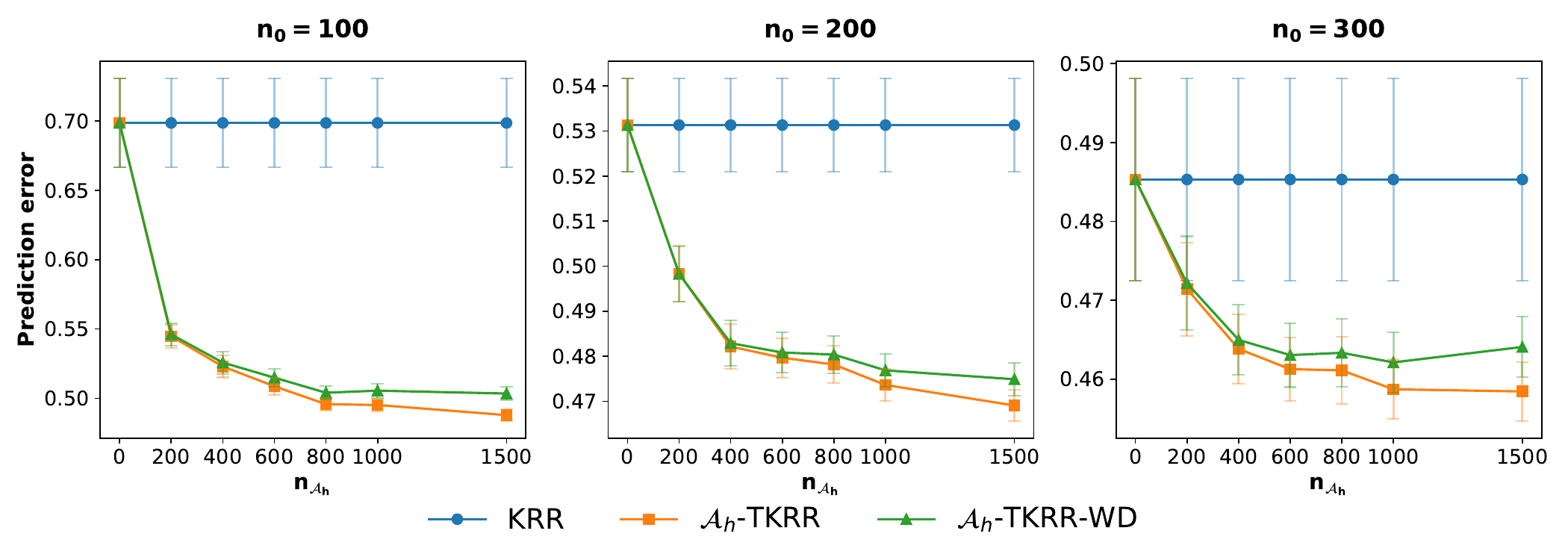}
    \caption{Averaged prediction errors of all the competitors under various scenarios in the wine quality data.}
\label{real_data_wine}
\end{figure}

In this application, ${\cal A}_h$ is known and only contains one source data, and thus
we apply ${\cal A}_h$-TKRR to this dataset and compare its performance to that of KRR and ${\cal A}_h$-TKRR-WD. Following the similar treatment as that in \cite{cai2022transfer}, we adopt random forest \citep{breiman2001random} to rank the feature importance and select the first three influential features: `alcohol', `sulfates', and `volatile acidity' for modeling. We are interested in the performance of the competitors in various scenarios with varying $n_0$ and $n_{{\cal A}_h}$. To be specific, we randomly select samples with size $n_{{\cal A}_h}$ and $n_0$ from the source and target data, respectively, and vary $n_{{\cal A}_h}$ from $\{0,200,400,600,800,1000,1500\}$ and  $n_0$ from $\{100,200,300\}$. We use all the remaining samples in the target data to evaluate the prediction accuracy. Each scenario is replicated 100 times and the averaged numerical performance is illustrated in Figure \ref{real_data_wine}.

From Figure \ref{real_data_wine}, we can conclude that ${\cal A}_h$-TKRR outperforms the other two methods in all the scenarios. This demonstrates the prediction performance of the red wine quality can be significantly improved by utilizing the source data. It is also interesting to notice that the prediction error of ${\cal A}_h$-TKRR decreases with the increasing of $n_{{\cal A}_h}$ and is much smaller than that of ${\cal A}_h$-TKRR-WD, which further verifies that the prediction enhancement is not simply due to the larger sample size, but also the effectiveness of transfer learning.

\subsection{UK used car data}\label{real:2}

In this subsection, we explore a practical application using the UK used car price dataset available at \url{https://www.kaggle.com/datasets/kukuroo3/used-car-price-dataset-competition-format}. The dataset contains data from nine prominent car brands in the UK market: Mercedes Benz (Merc), Ford, Volkswagen (VW), Bayerische Motoren Werke (BMW), Hyundai, Toyota, Skoda, Audi, and Vauxhall, with sample sizes of 1219, 1196, 1159, 978, 817, 717, 638, 523, and 385 respectively. The primary variable of interest is the price of each used car, with covariates that include essential details such as the registration year, mileage, road tax, miles per gallon (mpg), engine size, gearbox type, and fuel type. Our goal
is to predict used car prices for specific brands, thereby aiding sellers in determining the optimal time to sell. To this end, transfer learning is employed to improve the prediction accuracy for a target car brand by leveraging information from other brands.

For our analysis, we treat the car brand samples sequentially as the target data, using the remaining brands' samples as the source data. For each target dataset, we randomly select half of the samples for training and use the remaining half to assess prediction accuracy. In this context, we lack prior knowledge about which sources are transferable, so we adopt SA-TKRR. Additionally, we benchmark the performance of SA-TKRR against other methods discussed in Section \ref{5.2}, including KRR, Pooled-TKRR, AEW-TKRR, and D-TKRR. Each scenario is replicated 100 times, and the average numerical performance is presented in Table \ref{table1}.

\begin{table}[t]
\centering
\scriptsize
\caption{Prediction errors of all the competitors
in the UK used car data.
}
\vspace{0.1in}

\scalebox{1}{\setlength{\tabcolsep}{2mm}{
\begin{tabular}{ l c c c c c c c c c } 
\hline\hline
 \multirow{3}*{Methods} & \multicolumn{9}{c}{Target Brands}\\
\cmidrule(r){2-10}  &  Merc  & Ford    &  VW  & BMW  & Hyundai & Toyota & Skoda & Audi & Vauxhall\\
\hline
\multirow{2}*{SA-TKRR}  & \textbf{0.4918} & \textbf{0.0596}  &  0.1746 &\textbf{0.6686} & \textbf{0.0259}& 0.1211  & \textbf{0.0221} & \textbf{0.5364} & \textbf{0.0333} \\
  & \scriptsize{\textbf{(0.127)}} & \scriptsize{\textbf{(0.015)}} & \scriptsize{(0.014)} & \scriptsize{\textbf{(0.145)}}& \scriptsize{\textbf{(0.005)}}& \scriptsize{(0.024)}& \scriptsize{\textbf{(0.002)}}& \scriptsize{\textbf{(0.219)}}& \scriptsize{\textbf{(0.008)}}\\
  
\hline
\multirow{2}*{KRR}  & 0.5176 & 0.0623 & 0.1838 & 0.7179  & 0.0297 & 0.1562 & 0.0251 & 0.7327 & 0.0413\\
  & \scriptsize{(0.138) } & \scriptsize{(0.012) } & \scriptsize{(0.013) } & \scriptsize{(0.151) }& \scriptsize{(0.007) }& \scriptsize{(0.040) }& \scriptsize{(0.004) }& \scriptsize{(0.258) }& \scriptsize{(0.010) }\\
\hline
\multirow{2}*{Pooled-TKRR}  &{0.5109} & {0.0710}  & \textbf{0.1723}  &{0.6733}& 0.0371 & \textbf{0.1150} & {0.0230}& 0.5396 & 0.0488 \\
  & \scriptsize{{(0.112)} } & \scriptsize{{(0.026)} } & \scriptsize{\textbf{(0.013) }} & \scriptsize{{(0.150)} }& \scriptsize{(0.008) }& \scriptsize{\textbf{(0.015)} }& \scriptsize{{(0.002)} }& \scriptsize{(0.115) }& \scriptsize{(0.017) }\\



\hline

\multirow{2}*{AEW-TKRR}  & 0.5698 & 0.0687 & 0.1971  & 0.7962& 0.0386   & 0.1754 &  0.0246  & 0.7451 & 0.0532\\
 & \scriptsize{(0.123)} & \scriptsize{(0.014)} & \scriptsize{(0.018)} & \scriptsize{(0.185)}& \scriptsize{{(0.012)}}& \scriptsize{(0.036)}& \scriptsize{(0.002)}& \scriptsize{{(0.228)}}& \scriptsize{{(0.019)}}\\
 
\hline
\multirow{2}*{D-TKRR}  & 0.5130 & 0.0639  & 0.1778  & 0.6695& 0.0270    & 0.1169  & 0.0225  &0.5460 & 0.0342 \\
 & \scriptsize{(0.108 )} & \scriptsize{(0.023)} & \scriptsize{(0.014)} & \scriptsize{(0.152)}& \scriptsize{{(0.006)}}& \scriptsize{(0.020)}& \scriptsize{(0.002)}& \scriptsize{{(0.118)}}& \scriptsize{{(0.006)}}\\
  \hline  \hline
\end{tabular}
}}

 \label{table1}
\end{table}

From Table \ref{table1}, it is evident that SA-TKRR delivers superior prediction performance for all brands except VW and Toyota, highlighting the effectiveness of our proposed method. Notably, Pooled-TKRR demonstrates leading performance specifically for VW and Toyota, with SA-TKRR's performance being comparable to that of Pooled-TKRR. This can be largely attributed to VW and Toyota being popular brands with numerous best-selling models, which likely share more similarities with other brands, making the entire source data transferable. In contrast, other brands, which may exhibit lower similarities with the rest, do not benefit as much from pooling all source data. This is because the inclusion of non-relevant or negative data sources could detract from the overall prediction accuracy.

\section{Conclusion}\label{sec:con}
This paper conducts an in-depth investigation of the transfer learning problem associated with KRR from both methodological and theoretical standpoints. Its primary objective is to bridge the gap between practical effectiveness and theoretical guarantees, addressing the crucial question of whether the transfer estimation of KRR can achieve optimality, a topic of significant interest to the statistics and machine learning communities.
To address these challenges, the paper proposes two highly efficient transfer learning methods designed for scenarios where the transferable sources are either known or unknown. From a theoretical perspective, this study establishes the optimal minimax rate for the desired estimator in cases with known transferable sources, utilizing a range of techniques from empirical process theory to analyze kernel classes. Additionally, it demonstrates that the desired estimator in scenarios with unknown transferable sources can asymptotically approach a nearly optimal rate.
Notably, the paper also identifies several promising directions for future research. One particularly intriguing avenue for exploration involves a theoretical examination of the transfer learning problem in other kernel-based methods, such as kernel-based quantile regression and kernel support vector machines. 




\bibliographystyle{apalike}
\bibliography{sample}

\begin{thebibliography}{}

\bibitem[Bastani, 2021]{bastani2021predicting}
Bastani, H. (2021).
\newblock Predicting with proxies: Transfer learning in high dimension.
\newblock {\em Management Science}, \textbf{67}:2964--2984.

\bibitem[Blitzer et~al., 2007]{blitzer2007learning}
Blitzer, J., Crammer, K., Kulesza, A., Pereira, F., and Wortman, J. (2007).
\newblock Learning bounds for domain adaptation.
\newblock {\em Advances in Neural Information Processing Systems},
  \textbf{20}:129--136.

\bibitem[Breiman, 2001]{breiman2001random}
Breiman, L. (2001).
\newblock Random forests.
\newblock {\em Machine Learning}, \textbf{45}:5--32.

\bibitem[Cai and Pu, 2024]{cai2022transfer}
Cai, T.~T. and Pu, H. (2024).
\newblock Transfer learning for nonparametric regression: Non-asymptotic
  minimax analysis and adaptive procedure.
\newblock {\em arXiv:2401.12272}.

\bibitem[Cai and Wei, 2021]{cai2021transfer}
Cai, T.~T. and Wei, H. (2021).
\newblock Transfer learning for nonparametric classification: {M}inimax rate
  and adaptive classifier.
\newblock {\em The Annals of Statistics}, \textbf{49}(1):100--128.

\bibitem[Caponnetto and De~Vito, 2007]{caponnetto2007optimal}
Caponnetto, A. and De~Vito, E. (2007).
\newblock Optimal rates for the regularized least-squares algorithm.
\newblock {\em Foundations of Computational Mathematics}, \textbf{7}:331--368.

\bibitem[Cortez et~al., 2009]{cortez2009modeling}
Cortez, P., Cerdeira, A., Almeida, F., Matos, T., and Reis, J. (2009).
\newblock Modeling wine preferences by data mining from physicochemical
  properties.
\newblock {\em Decision Support Systems}, \textbf{47}(4):547--553.

\bibitem[Do and Ng, 2005]{do2005transfer}
Do, C.~B. and Ng, A.~Y. (2005).
\newblock Transfer learning for text classification.
\newblock {\em Advances in Neural Information Processing Systems 18}, pages
  299--306.

\bibitem[Feng et~al., 2023]{feng2023towards}
Feng, X., He, X., Wang, C., Wang, C., and Zhang, J. (2023).
\newblock Towards a unified analysis of kernel-based methods under covariate
  shift.
\newblock {\em Advances in Neural Information Processing Systems 36}, pages
  73839--73851.

\bibitem[Fischer and Steinwart, 2020]{fischer2020sobolev}
Fischer, S. and Steinwart, I. (2020).
\newblock Sobolev norm learning rates for regularized least-squares algorithms.
\newblock {\em Journal of Machine Learning Research}, \textbf{21}(205):1--38.

\bibitem[Ga{\^\i}ffas and Lecu{\'e}, 2011]{gaiffas2011hyper}
Ga{\^\i}ffas, S. and Lecu{\'e}, G. (2011).
\newblock Hyper-sparse optimal aggregation.
\newblock {\em Journal of Machine Learning Research}, \textbf{12}:1813--1833.

\bibitem[Garg et~al., 2020]{garg2020unified}
Garg, S., Wu, Y., Balakrishnan, S., and Lipton, Z. (2020).
\newblock A unified view of label shift estimation.
\newblock {\em Advances in Neural Information Processing Systems 33}, pages
  3290--3300.

\bibitem[Ge et~al., 2014]{ge2014handling}
Ge, L., Gao, J., Ngo, H., Li, K., and Zhang, A. (2014).
\newblock On handling negative transfer and imbalanced distributions in
  multiple source transfer learning.
\newblock {\em Statistical Analysis and Data Mining}, \textbf{7}:254--271.

\bibitem[He et~al., 2021]{he2021efficient}
He, X., Wang, J., and Lv, S. (2021).
\newblock Efficient kernel-based variable selection with sparsistency.
\newblock {\em Statistica Sinica}, \textbf{31}(4):2123--2151.

\bibitem[Kimeldorf and Wahba, 1971]{kimeldorf1971some}
Kimeldorf, G. and Wahba, G. (1971).
\newblock Some results on tchebycheffian spline functions.
\newblock {\em Journal of Mathematical Analysis and Applications},
  \textbf{33}(1):82--95.

\bibitem[Koltchinskii and Yuan, 2010]{Koltchinskii2010SPARSITYIM}
Koltchinskii, V. and Yuan, M. (2010).
\newblock Sparsity in multiple kernel learning.
\newblock {\em The Annals of Statistics}, \textbf{38}:3660--3695.

\bibitem[Lee et~al., 2025]{lee2025doubly}
Lee, S.-h., Ma, Y., and Zhao, J. (2025).
\newblock Doubly flexible estimation under label shift.
\newblock {\em Journal of the American Statistical Association},
  120(549):278--290.

\bibitem[Leung and Barron, 2006]{leung2006information}
Leung, G. and Barron, A.~R. (2006).
\newblock Information theory and mixing least-squares regressions.
\newblock {\em IEEE Transactions on Information Theory},
  \textbf{52}(8):3396--3410.

\bibitem[Li et~al., 2022a]{li2022transfera}
Li, S., Cai, T.~T., and Li, H. (2022a).
\newblock Transfer learning for high-dimensional linear regression: Prediction,
  estimation and minimax optimality.
\newblock {\em Journal of the Royal Statistical Society Series B: Statistical
  Methodology}, \textbf{84}(1):149--173.

\bibitem[Li et~al., 2022b]{li2022transferb}
Li, S., Cai, T.~T., and Li, H. (2022b).
\newblock Transfer learning in large-scale gaussian graphical models with false
  discovery rate control.
\newblock {\em Journal of the American Statistical Association}, pages 1--13.

\bibitem[Li et~al., 2024]{li2023estimation}
Li, S., Zhang, L., Cai, T.~T., and Li, H. (2024).
\newblock Estimation and inference for high-dimensional generalized linear
  models with knowledge transfer.
\newblock {\em Journal of the American Statistical Association},
  \textbf{119}:1274--1285.

\bibitem[Lin and Reimherr, 2022]{lin2022transfer}
Lin, H. and Reimherr, M. (2022).
\newblock Transfer learning for functional linear regression with structural
  interpretability.
\newblock {\em arXiv preprint arXiv:2206.04277}.

\bibitem[Lin et~al., 2020]{lin2020distributed}
Lin, S.-B., Wang, D., and Zhou, D.-X. (2020).
\newblock Distributed kernel ridge regression with communications.
\newblock {\em Journal of Machine Learning Research}, \textbf{21}:1--38.

\bibitem[Ma et~al., 2023]{ma2023optimally}
Ma, C., Pathak, R., and Wainwright, M.~J. (2023).
\newblock Optimally tackling covariate shift in rkhs-based nonparametric
  regression.
\newblock {\em The Annals of Statistics}, \textbf{51}(2):738--761.

\bibitem[Mansour et~al., 2009]{mansour2009domain}
Mansour, Y., Mohri, M., and Rostamizadeh, A. (2009).
\newblock Domain adaptation: Learning bounds and algorithms.
\newblock {\em The 22nd Annual Conference on Learning Theory}.

\bibitem[Murphy, 2012]{murphy2012machine}
Murphy, K.~P. (2012).
\newblock {\em Machine Learning: A Probabilistic Perspective}.
\newblock MIT Press.

\bibitem[Pan and Yang, 2010]{pan2010survey}
Pan, S.~J. and Yang, Q. (2010).
\newblock A survey on transfer learning.
\newblock {\em IEEE Transactions on Knowledge and Data Engineering},
  \textbf{22}:1345--1359.

\bibitem[Pathak et~al., 2022]{pathak2022new}
Pathak, R., Ma, C., and Wainwright, M. (2022).
\newblock A new similarity measure for covariate shift with applications to
  nonparametric regression.
\newblock In {\em Proceedings of the 39th International Conference on Machine
  Learning}, pages 17517--17530.

\bibitem[Raghu et~al., 2019]{raghu2019transfusion}
Raghu, M., Zhang, C., Kleinberg, J., and Bengio, S. (2019).
\newblock Transfusion: Understanding transfer learning for medical imaging.
\newblock {\em Advances in Neural Information Processing Systems 32}, pages
  3347--3357.

\bibitem[Sch{\"o}lkopf and Smola, 2002]{scholkopf2002learning}
Sch{\"o}lkopf, B. and Smola, A.~J. (2002).
\newblock {\em Learning with {K}ernels: {S}upport {V}ector {M}achines,
  {R}egularization, {O}ptimization, and {B}eyond}.
\newblock MIT press.

\bibitem[Seah et~al., 2012]{seah2012combating}
Seah, C.-W., Ong, Y.-S., and Tsang, I.~W. (2012).
\newblock Combating negative transfer from predictive distribution differences.
\newblock {\em IEEE Transactions on Cybernetics}, \textbf{43}:1153--1165.

\bibitem[Shimodaira, 2000]{shimodaira2000improving}
Shimodaira, H. (2000).
\newblock Improving predictive inference under covariate shift by weighting the
  log-likelihood function.
\newblock {\em Journal of Statistical Planning and Inference},
  \textbf{90}(2):227--244.

\bibitem[Smale and Zhou, 2007]{smale2007learning}
Smale, S. and Zhou, D.-X. (2007).
\newblock Learning theory estimates via integral operators and their
  approximations.
\newblock {\em Constructive Approximation}, \textbf{26}(2):153--172.

\bibitem[Sugiyama and M{\"u}ller, 2005]{sugiyama2005model}
Sugiyama, M. and M{\"u}ller, K.-R. (2005).
\newblock Model selection under covariate shift.
\newblock In {\em Proceedings of the 15th International Conference on
  Artificial Neural Networks: Formal Models and Their Applications-Volume Part
  II}, pages 235--240.

\bibitem[Taylor and Stone, 2009]{taylor2009transfer}
Taylor, M.~E. and Stone, P. (2009).
\newblock Transfer learning for reinforcement learning domains: A survey.
\newblock {\em Journal of Machine Learning Research}, \textbf{10}:1633--1685.

\bibitem[Tian and Feng, 2023]{tian2022transfer}
Tian, Y. and Feng, Y. (2023).
\newblock Transfer learning under high-dimensional generalized linear models.
\newblock {\em Journal of the American Statistical Association},
  \textbf{118}(544):2684--2697.

\bibitem[Torrey and Shavlik, 2010]{torrey2010transfer}
Torrey, L. and Shavlik, J. (2010).
\newblock Transfer learning.
\newblock In {\em Handbook of {R}esearch on {M}achine {L}earning {A}pplications
  and {T}rends: {A}lgorithms, {M}ethods, and {T}echniques}, pages 242--264. IGI
  Global.

\bibitem[Vijayakumar et~al., 2002]{vijayakumar2002statistical}
Vijayakumar, S., D'souza, A., Shibata, T., Conradt, J., and Schaal, S. (2002).
\newblock Statistical learning for humanoid robots.
\newblock {\em Autonomous Robots}, \textbf{12}:55--69.

\bibitem[Wahba, 1990]{wahba1990spline}
Wahba, G. (1990).
\newblock Spline models for observational data.
\newblock In {\em CBMS-NSF Regional Conference Series in Applied Mathematics}.
  SIAM.

\bibitem[Wang et~al., 2016]{wang2016nonparametric}
Wang, X., Oliva, J.~B., Schneider, J., and P{\'o}czos, B. (2016).
\newblock Nonparametric risk and stability analysis for multi-task learning
  problems.
\newblock In {\em Proceedings of the 25-th International Joint Conference on
  Artificial Intelligence}, pages 2146--2152.

\bibitem[Weiss et~al., 2016]{weiss2016survey}
Weiss, K., Khoshgoftaar, T.~M., and Wang, D. (2016).
\newblock A survey of transfer learning.
\newblock {\em Journal of Big Data}, \textbf{3}:1--40.

\bibitem[Zhang et~al., 2023]{zhang2023optimality}
Zhang, H., Li, Y., Lu, W., and Lin, Q. (2023).
\newblock On the optimality of misspecified kernel ridge regression.
\newblock In {\em International Conference on Machine Learning}, pages
  41331--41353. PMLR.

\bibitem[Zhang et~al., 2014]{zhang2014spectral}
Zhang, Y., Chen, X., Zhou, D., and Jordan, M.~I. (2014).
\newblock Spectral methods meet {EM}: {A} provably optimal algorithm for
  crowdsourcing.
\newblock {\em Advances in Neural Information Processing Systems},
  \textbf{27}:1260--1268.

\bibitem[Zhang and Zhu, 2025]{zhang2022transfer}
Zhang, Y. and Zhu, Z. (2025).
\newblock Transfer learning for high-dimensional quantile regression via
  convolution smoothing.
\newblock {\em Statistica Sinica}, \textbf{35}:1--39.

\end{thebibliography}

\end{document}